\documentclass{article} 
\usepackage{iclr2026_conference,times}
\usepackage{multirow}
\usepackage{multicol} 
\usepackage{array}    
\usepackage{amssymb}  
\usepackage{pifont}   
\usepackage{tabularx}
\usepackage[utf8]{inputenc}
\usepackage{caption}  
\usepackage{subcaption} 
\usepackage[table]{xcolor}
\usepackage{booktabs}
\usepackage{adjustbox} 
\usepackage{rotating}

\usepackage{amsmath,amsfonts,bm}

\def\eqref#1{equation~\ref{#1}}

\def\1{\bm{1}}

\DeclareMathAlphabet{\mathsfit}{\encodingdefault}{\sfdefault}{m}{sl}
\SetMathAlphabet{\mathsfit}{bold}{\encodingdefault}{\sfdefault}{bx}{n}

\usepackage{hyperref}
\usepackage{url}

\title{Unveiling Deep Semantic Uncertainty Perception for Language-Anchored Multi-modal Vision-Brain Alignment}

\author{Zehui Feng$^{1}$, Chenqi Zhang$^{1}$, Mingru Wang$^{1}$, Minuo Wei$^{1}$, Shiwei Cheng$^{4}$, \\
\textbf{Cuntai Guan$^{2,*}$, Ting Han$^{1,3,}$} \thanks{Corresponding authors.} \\
\\
$^{1}$Shanghai Jiao Tong University, Shanghai, China\\
\\
$^{2}$Nanyang Technological University, Singapore\\
\\
$^{3}$Zhejiang University, Hangzhou, China\\
\\
$^{4}$Zhejiang University of Technology, Hangzhou, China\\
\\
\{fzh$\_$sjtu, zhangchenqi1101, awmruc, minuo1029\}@sjtu.edu.cn, \\
swc@zjut.edu.cn, ctguan@ntu.edu.sg, hanting@sjtu.edu.cn  \\
\\
\textcolor{red}{\texttt{\url{https://github.com/DanceSkyCode/Bratrix}}}
}
\iclrfinalcopy 
\begin{document}

\maketitle
\begin{abstract}
Unveiling visual semantics from neural signals such as EEG, MEG, and fMRI remains a fundamental challenge due to subject variability and the entangled nature of visual features. Existing approaches primarily align neural activity directly with visual embeddings, but visual-only representations often fail to capture latent semantic dimensions, limiting interpretability and deep robustness. To address these limitations, we propose \textbf{Bratrix}, the first end-to-end framework to achieve multimodal Language-Anchored Vision–Brain alignment. Bratrix decouples visual stimuli into hierarchical visual and linguistic semantic components, and projects both visual and brain representations into a shared latent space, enabling the formation of aligned visual–language and brain–language embeddings. To emulate human-like perceptual reliability and handle noisy neural signals, Bratrix incorporates a novel uncertainty perception module that applies uncertainty-aware weighting during alignment. By leveraging learnable language-anchored semantic matrices to enhance cross-modal correlations and employing a two-stage training strategy of single-modality pretraining followed by multimodal fine-tuning, \textbf{Bratrix-M} improves alignment precision. Extensive experiments on EEG, MEG, and fMRI benchmarks demonstrate that Bratrix improves retrieval, reconstruction, and captioning performance compared to state-of-the-art methods, specifically surpassing \textbf{14.3\%} in 200-way EEG retrieval task. Code and model are available.
\end{abstract}
\begin{figure}[h]
    \centering
    \includegraphics[width=0.7\linewidth]{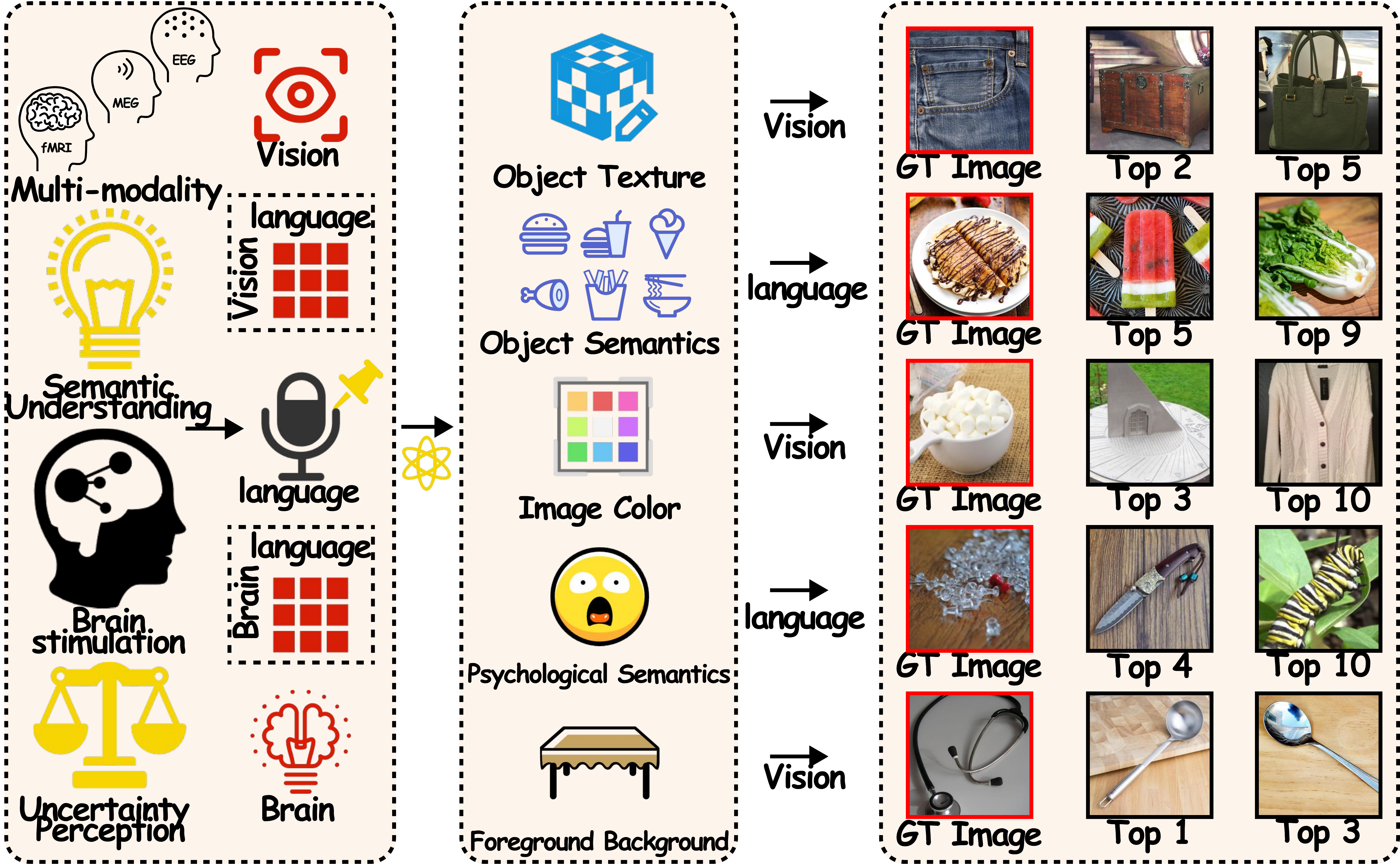}
    \caption{Overall of Language-Anchored Vision-Brain Alignment. }
    \label{fig:0}
\end{figure}

\section{Introduction}
\label{sec:intro}
Unveiling how the human brain understands visual representations has been a central challenge in cognitive neuroscience and pattern recognition application. Neural modalities such as EEG \cite{song2024decodingnaturalimageseeg, liu2025vieeghierarchicalvisualneural, zhang2024cognitioncapturerdecodingvisualstimuli}, MEG \cite{11094846, li2024visualdecodingreconstructioneeg, benchetrit2024braindecodingrealtimereconstruction}, and fMRI \cite{517004, scotti2024mindeye2sharedsubjectmodelsenable, scotti2023reconstructingmindseyefmritoimage, chen2023seeingbrainconditionaldiffusion} offer complementary insights into the temporal and spatial dynamics of brain activity, yet decoding such signals into meaningful semantic representations remains highly challenging due to subject variability \cite{chen2024visualneuraldecodingimproved}, and the intrinsic complexity \cite{NEURIPS2024_84bad835, FENG2025103363} of neural responses. 

Existing studies on brain–vision alignment primarily focus on aligning neural activity directly with visual feature spaces \cite{11104249}. However, visual representations themselves are not free from limitations: they often lack deep implicit disentanglement of semantic dimensions (e.g., psychological semantics, object semantics, and contextual environment relationships) \cite{Wilson2023}, and are particularly fragile when neural signals are noisy or incomplete. Relying solely on vision-based representations can lead to partial or biased decoding results \cite{Fu_2025,bai2023dreamdiffusiongeneratinghighqualityimages}. At the same time, large-scale pretrained models in computer vision and natural language processing have demonstrated remarkable capabilities in capturing structured semantic spaces, offering new opportunities to bridge the gap between raw neural signals and high-level perceptual semantics.

To address these challenges, we propose \textbf{Bratrix}, a novel framework that anchors brain and vision alignment in language semantics. As the examples shown in Fig. \ref{fig:0}, Language provides a structured and compositional representation that captures richer and more disentangled information than visual embeddings alone. Bratrix first extracts deep hierarchical visual representations through Vision Semantic Decoupling module and generates fine-grained linguistic anchors in Language Semantic Decoupling module. Both of them capture the structured semantics of stimuli across modalities. Considering the noise in neural recordings, Bratrix introduces a Vision-Language Semantic Uncertainty perception mechanism that emulates the human ability to estimate the reliability of perceptual signals. These representations are then integrated to construct \textbf{language-anchored visual and brain semantic matrices}, and the visual and brain representations are further encoded into a shared latent space to produce unified modality-specific embeddings. By progressively aligning these semantic matrices via decoder representations from latent space, Bratrix achieves rich language-semantics-anchored cross-modal correspondence, which disentangles semantic components, enhance potential correlations, preserves fine-grained perceptual details, and enhances robustness across subjects and recording modalities. Beyond these architectural contributions, Bratrix adopts a two-stage training strategy. By first leveraging single-modality pretraining to establish strong unimodal priors, it then performs multi-modal fine-tuning, yielding \textbf{Bratrix-M}. Both of Matrix and Matrix-M achieves significantly higher alignment precision across EEG, MEG, and fMRI. Extensive evaluations including retrieval, reconstruction, captioning, ablation studies, and visualization analyses, show that Bratrix consistently surpasses state-of-the-art baselines, achieving unprecedented performance in decoding and aligning brain activity with rich semantic content.  

In summary, our contributions are threefold:  
\begin{itemize}
    \item We propose \textbf{Bratrix}, the first end-to-end framework to explicitly leverage language semantics as a semantic anchor for brain–vision alignment, fully exploiting the latent semantic relationships between neural signals and visual representations.
    \item We propose a novel uncertainty perception module that quantifies the reliability of neural signals in a human-inspired pattern, enabling Bratrix to incorporate uncertainty-aware weighting during feature alignment and thereby enhance cross-modal semantic decoding.
    \item We adopt a two-stage training strategy that first pretrains on single-modality representations and then fine-tunes across multiple modalities, fully leveraging the semantic matrices of different modalities. Our results show that this multimodal joint training significantly improves alignment performance.
\end{itemize}
\section{Related Work}
\label{sec:formatting}

\subsection{Multimodal Brain Signal Visual Decoding}
Recent years have obtained rapid progress in decoding visual information from single-modality brain signals such as fMRI~\cite{wang2024mindbridgecrosssubjectbraindecoding, scotti2023reconstructingmindseyefmritoimage, huo2024neuropictorrefiningfmritoimagereconstruction, Allen2022} or EEG~\cite{li2024visualdecodingreconstructioneeg, 3681292, li2025neuralmcrlneuralmultimodalcontrastive, liu2025vieeghierarchicalvisualneural, wu2025bridgingvisionbraingapuncertaintyaware, song2024decodingnaturalimageseeg}. fMRI-based approaches benefit from high spatial resolution, enabling implicit capture of the hierarchical organization of visual processing from early edge detection to complex semantic representation~\cite{huo2024neuropictorrefiningfmritoimagereconstruction}. Conversely, EEG provides millisecond-level temporal precision but suffers from coarse spatial resolution \cite{8257015, 8099962, ZHENG2021102174}, often leading to flattened feature representations that fail to reflect the brain’s progressive refinement from low- to high-level features ~\cite{wu2025bridgingvisionbraingapuncertaintyaware}. While single-modality decoding has achieved remarkable performance in image reconstruction~\cite{li2024visualdecodingreconstructioneeg}, image retrieval~\cite{3681292, chen2024visualneuraldecodingimproved}, image captioning~\cite{li2025brainflorauncoveringbrainconcept}, and video reconstruction~\cite{NEURIPS2024_84bad835}, the inherent limitations of each modality restrict generalization across subjects and tasks. Multimodal decoding jointly leverages complementary neural recordings, and offers the potential to integrate spatial and temporal advantages. However, existing multimodal approaches ~\cite{li2024visualdecodingreconstructioneeg, gao2025cinebrainlargescalemultimodalbrain} often fail to construct a unified latent feature space in which different modalities are non-exclusive and mutually complementary. Instead, modality-specific features tend to dominate independently, limiting the ability of one modality to enhance the performance of another.

\subsection{Cross-Modal Alignment in Brain and Vision}
Recent brain vision alignment methods typically project temporal features into a visual embedding ~\cite{radford2021learningtransferablevisualmodels, song2024decodingnaturalimageseeg} via contrastive objectives learning, enabling zero-shot retrieval or reconstruction. However, EEG does not function as a global feature fuser of visual content~\cite{Lawhern_2018, li2024visualdecodingreconstructioneeg, 3681292, li2025neuralmcrlneuralmultimodalcontrastive}; rather, under limited capacity and task demands, it selectively amplifies the most diagnostic cues among multiple, potentially competing semantic signals, such as color, contour ~\cite{liu2025vieeghierarchicalvisualneural}, layout~\cite{zhang2024cognitioncapturerdecodingvisualstimuli}, semantic~\cite{li2025neuralmcrlneuralmultimodalcontrastive}, or category-level evidence. This selectivity is consistent with electrophysiological findings ~\cite{Yves2023, HAN2019125} that feature-based attention enhances responses to attended attributes (e.g., color or orientation)~\cite{zhang2024cognitioncapturerdecodingvisualstimuli} while leaving unattended attributes underrepresented, and with perceptual phenomena where observers experience feature awareness without object identity due to incomplete awareness. Consequently, contrastive alignment that assumes a single, uniformly semantic EEG representation risks overfitting to whichever cue is most discriminative in the training distribution, collapsing hierarchical information and diminishing cross-modal complementarity~\cite{liu2025vieeghierarchicalvisualneural}. A more faithful alignment should preserve the natural competition and hierarchy among cues, enabling EEG representations to convey graded evidence from low- to high-level features, rather than collapsing them into a single undifferentiated semantic representation.
\section{Proposed Model}
\label{sec:proposed_model}
\begin{figure*}[t]
    \centering
    \includegraphics[width=1\linewidth]{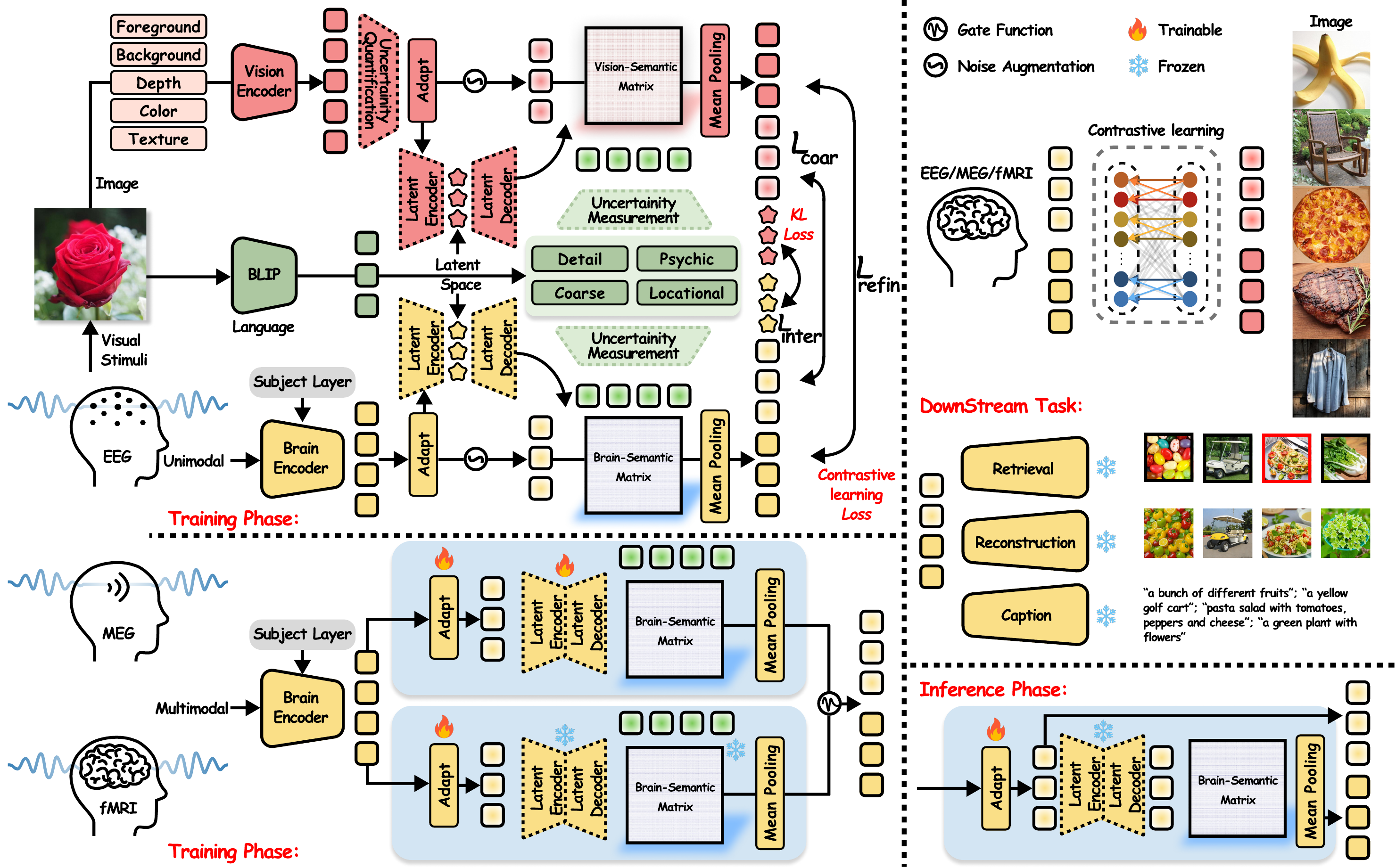}
    \caption{Overall Framework of Bratrix. Bratrix comprises Vision semantic decoupling module, a Brain encoder module, a language semantic decoupling module, and a language-anchored visual-brain alignment module. There are totally four stages in this framework: single-modal pre-training phase, multi-modal fine-tuning phase, inference phase, and downstream task phase.}
    \label{fig:overall}
\end{figure*}
\textbf{Formulation. }Let the neural recordings be denoted as $\mathbf{X} \in \mathbb{R}^{N \times C \times T}$, where $N$ is the number of samples, $C$ the channel dimension, and $T$ the temporal length (omitted for fMRI). Our objective is to learn neural embeddings $\mathbf{Z} = E(\mathbf{X}) \in \mathbb{R}^{N \times F}$ through a unified projector $E$, with $F$ the embedding dimension. In parallel, we obtain image embeddings $\hat{\mathbf{Z}} \in \mathbb{R}^{N \times F}$ from the CLIP encoder given stimuli $\mathbf{I}$. The training goal is to align neural and visual embeddings, enabling cross-modal correspondence.

\textbf{Overall Architecture. }We propose an end-to-end framework for aligning and enhancing multi-modal brain neural data, motivated by the brain’s ability to integrate sensory information from different modalities into higher-level abstract representations that support reasoning and communication. In our design, language serves as the central semantic anchor, providing a symbolic and expressive space for unifying heterogeneous neural and visual signals. At the core of the framework is the language-anchored brain–vision semantic matrix, which explicitly leverages language grounding to facilitate efficient cross-modal alignment. To further capture the variability of neural responses, we introduce a semantic uncertainty perception module that preserves the fine-grained specificity of subjects’ reactions to visual stimuli, including perceptual attributes such as color, texture, spatial layout, and affective impressions. Training proceeds under a contrastive learning paradigm that jointly optimizes brain and vision encoders, coupled with uncertainty-aware semantic modulation within a shared latent space. In addition, a pre-trained semantic encoder is integrated with frozen parameters, while its modulation and interaction layers remain shared across modalities, enabling the reuse of linguistic priors. During inference, the unified embeddings derived from neural signals generalize seamlessly to zero-shot applications, including EEG/MEG/fMRI-based image classification, retrieval, and reconstruction. The overall architecture is illustrated in Fig.~\ref{fig:overall}.

\subsection{Vision Semantic Decoupling.}
Human visual perception is inherently hierarchical, influenced by both cognitive processes and temporal dynamics. Motivated by this, we decompose visual stimuli into different levels of semantic layers in perceptual granularity: foreground, background, spatial layout, color, and texture. Pre-trained vision encoders (see Appendix) are employed to extract the corresponding representations $\{\mathbf{V}_f, \mathbf{V}_b, \mathbf{V}_d, \mathbf{V}_c, \mathbf{V}_t\}$. 
\subsection{Language-Anchored Vision-Brain Alignment.}
\textbf{Language Semantic Decoupling: }Language encodes abstract and compositional semantics, making it an ideal anchor for aligning heterogeneous modalities. To leverage this property, we incorporate language as an auxiliary semantic anchor, projecting textual embeddings into both visual and neural feature spaces. Specifically, we employ the BLIP model to generate multi-level textual semantics $\{\mathbf{T}_c, \mathbf{T}_d, \mathbf{T}_l, \mathbf{T}_p\}$, including coarse captions, detail descriptions, locational attributes, and Psychic impressions, which guide the alignment of vision and brain signals around a shared semantic reference. More details of textual semantics are shown in Appendix. Visual features are grounded in these linguistic semantics, while neural signals, modeled by an iTransformer-based encoder, are mapped to the same space to capture temporal dynamics and cognitive variability.

\textbf{Visual-Language Semantic Uncertainty Perception: }
Considering the reliability across semantic components of visual and textual features, we introduce a Semantic Uncertainty Perception (SUP) module. Given visual features $\mathbf{V} \in \mathbb{R}^{B \times C_v \times K}$ and textual features $\mathbf{T} \in \mathbb{R}^{B \times C_T \times K}$, we first estimate semantic evidence \cite{CHAN200567} into Dirichlet parameters via learnable heads:
\begin{equation}
\boldsymbol{\alpha}_V = \mathrm{exp}(\xi(\mathbf{V})) + 1,
\boldsymbol{\alpha}_T = \mathrm{exp}(\xi(\mathbf{T})) + 1,
\end{equation}
where $\mathrm{exp}$ denotes the exponential function, $\xi$ denotes the FFN layer, $B$ is the batch size, $C$ is the number of semantic components, and $K$ is the dimension of features. The uncertainty of each component is computed as
\begin{equation}
u_V^c = \frac{K}{\sum_{k=1}^{K} \alpha_V^{c,k}},
u_T^c = \frac{K}{\sum_{k=1}^{K} \alpha_T^{c,k}},c = \{1, \dots, C\},
\end{equation}
and transformed into reliability weights $w = 1 - u$. Then, we perform a weighted fusion across semantic components with normalization:
\begin{equation}
\tilde{\mathbf{V}} = \xi(\frac{\sum_{c=1}^{C} w_V^c \mathbf{V}^c}{\sum_{c=1}^{C} w_V^c}),
\tilde{\mathbf{T}} = \xi(\frac{\sum_{c=1}^{C} w_T^c \mathbf{T}^c}{\sum_{c=1}^{C} w_T^c})
\end{equation}

\textbf{Language-Anchored Visual–Brain Semantic Alignment:}  
A key challenge in cross-modal alignment lies in the noisy and unconstrained interactions obtained by directly multiplying brain and visual features with language embeddings. To address this challenge, we introduce a language-anchored semantic alignment module that introduces structured semantic matrices, learnable priors, and a regularization loss to ensure stable and interpretable alignment in the latent space. Mathematically, the brain neural signal $\tilde{\mathbf{B}}$ is encoded via iTransformer \cite{liu2024itransformerinvertedtransformerseffective}. We first compute modality–language interaction matrices:
\begin{equation}
\hat{\mathbf{M}}_B = \phi(\tilde{\mathbf{B}})\phi(\tilde{\mathbf{T}})^\top,
\hat{\mathbf{M}}_V = \phi(\tilde{\mathbf{V}})\phi(\tilde{\mathbf{B}})^\top,
\end{equation}
where $\phi(\cdot)$ is a learnable linear projection. To encourage cross-modal consistency, we align the distributions of sparse representations extracted from brain and vision features via a symmetric KL divergence \cite{cui2025generalizedkullbackleiblerdivergenceloss}:
\begin{equation}
\breve{\mathbf{M}}_B=\Upsilon(\hat{\mathbf{M}}_B),
\breve{\mathbf{M}}_V=\Upsilon(\hat{\mathbf{M}}_V),
\end{equation}
\begin{equation}
\mathcal{L}_{\text{align}} = D_{\mathrm{KL}}(\breve{\mathbf{M}}_B \| \breve{\mathbf{M}}_V) + D_{\mathrm{KL}}(\breve{\mathbf{M}}_V\|\breve{\mathbf{M}}_B),
\end{equation}
where $\Upsilon$ denotes the latent encoder, $\breve{\mathbf{M}}_B$ and $\breve{\mathbf{M}}_V$ are shared latent representation. However, direct interactions in eq. (4) may entangle spurious correlations. To refine them, we build a lightweight re-weight function and pooled the matrices from sparse representations and prior structures:
\begin{equation}
\tilde{\mathbf{M}}_B = \mathbf{P}(\breve{\mathbf{M}}_B\cdot\sigma(\partial(\breve{\mathbf{M}}_B) + \mathbf{P}_B)),
\end{equation}
\begin{equation}
\tilde{\mathbf{M}}_V =\mathbf{P}(\breve{\mathbf{M}}_V\cdot\sigma(\partial(\breve{\mathbf{M}}_V) + \mathbf{P}_V)),
\end{equation}
where $\partial$ denotes the latent decoder, $\mathbf{P}_B,\mathbf{P}_V \in \mathbb{R}^{K \times K}$ are learnable priors, $\mathbf{P}$ is average pooling function, and $\sigma(\cdot)$ is the sigmoid function. Finally, language-aligned embeddings are concatenated with the initial features to form unified representations.
\subsection{Contrastive Loss Function}  
To enforce semantic consistency between brain and visual modalities, we employ a contrastive learning \cite{NEURIPS2020_d89a66c7} framework. We compute cosine similarities between paired embeddings $\mathbf{F}_{\text{brain}}=[\tilde{\mathbf{V}};\tilde{\mathbf{M}}_V];\mathbf{F}_{\text{vis}}=[\tilde{\mathbf{B}};\tilde{\mathbf{M}}_B]$, scaled by a learnable temperature parameter $\alpha = \exp(\text{logit\_scale})$. The final objective is the cross-entropy loss over similarity logits:  
\begin{equation}
\mathcal{L}_{\text{con}} = \text{CrossEntropy}\!\left(\cos(\mathbf{F}_{\text{brain}}, \mathbf{F}_{\text{vis}}), Y\right),
\end{equation}
\begin{equation}
\mathcal{L}_{\text{total}} =\mathcal{L}_{\text{con}} +\alpha \cdot \mathcal{L}_{\text{align}},
\end{equation}
where $Y$ denotes the ground-truth labels, and $\alpha$ denotes the balance parameter. This loss encourages brain representations to align closely with their visual counterparts in a shared embedding space. Moreover, the aligned embeddings naturally enable three downstream tasks, which followed: (i) retrieval via similarity scores \cite{li2025brainflorauncoveringbrainconcept}, (ii) reconstruction \cite{scotti2024mindeye2sharedsubjectmodelsenable, hu2021loralowrankadaptationlarge}, and (iii) caption generation \cite{song2024decodingnaturalimageseeg, awadalla2023openflamingoopensourceframeworktraining} by fine-tuning a pretrained multi-modal LLM.
\subsection{Multi-modal Joint Training Strategy}  
To achieve unified representation across heterogeneous brain signals, we adopt a joint training strategy under the assumption of shared latent semantics across modalities. Specifically, we pretrained the Language-Anchored Visual-Brain Semantic Alignment module on one brain signals, which effectively approximates rich image-level semantic embeddings in brain embeddings. When incorporating additional modalities such as MEG or fMRI, we froze the pretrained SUP and SLA modules, and trained an adaptive linear adapter at the input layer. A gating function balances frozen semantic priors with trainable modality-specific parameters, enabling efficient transfer and robust alignment across diverse neural signals.
\begin{table*}[t]
\centering
\scriptsize
\caption{Image retrieval comparison Performance of Bratrix and Bratrix-M on THINGS-EEG2, THINGS-MEG, and THINGS-fMRI datasets under subject-dependent and subject-independent settings. Note that Bratrix-M represents the multi-modal fine-tuning model.}
\renewcommand{\arraystretch}{1} 
\setlength{\tabcolsep}{3pt}
\resizebox{\textwidth}{!}{%
\begin{tabular}{lcccccc|ccccc|ccccc}
\hline
\multirow{3}{*}{Model} & \multirow{3}{*}{Paper} 
& \multicolumn{5}{c}{\textbf{THINGS-EEG2}} 
& \multicolumn{5}{c}{\textbf{THINGS-MEG}} 
& \multicolumn{5}{c}{\textbf{THINGS-fMRI}} \\
\cline{3-17}
&& \shortstack{2-way \\ Top-1} & \shortstack{4-way \\ Top-1} & \shortstack{50-way \\ Top-1} & \shortstack{200-way \\ Top-1} & \shortstack{200-way \\ Top-5}
& \shortstack{2-way \\ Top-1} & \shortstack{4-way \\ Top-1} & \shortstack{10-way \\ Top-1} & \shortstack{200-way \\ Top-1} & \shortstack{200-way \\ Top-5}
& \shortstack{2-way \\ Top-1} & \shortstack{4-way \\ Top-1} & \shortstack{10-way \\ Top-1} & \shortstack{100-way \\ Top-1} & \shortstack{100-way \\ Top-5} \\
\hline
\multicolumn{16}{c}{\textbf{Subject-dependent}} \\ 
\hline
LSTM \cite{vennerød2021longshorttermmemoryrnn}& /& 90.6& 60.6& 40.7& 14.6& 21.0& 80.9 & 49.7& 10.8& 9.4& 18.6& 88.3& 49.4& 14.6& 9.4& 23.0\\
ConvNet \cite{radford2021learningtransferablevisualmodels}& /&91.3& 64.5& 43.5& 13.4& 19.1& 82.3& 52.0& 13.0& 11.1& 20.3& 89.2& 52.5& 17.0& 12.8& 25.0\\
EEGNet \cite{Lawhern_2018}& /&92.6& 67.9& 46.7& 16.2& 22.4& 81.9& 54.4& 14.4& 13.0& 23.9& 89.0& 54.8& 18.6& 14.9& 29.5\\
\midrule
MindEyev2 \cite{scotti2024mindeye2sharedsubjectmodelsenable}& ICML'2024& 93.2& 81.2& 65.0& 27.5& 59.5& 83.6& 54.7& 15.3& 13.5& 49.8& 90.6& 55.0& 19.4& 15.0& 58.2\\
BraVL \cite{du2023decodingvisualneuralrepresentations}& TPAMI'2023& 94.0& 78.8& 63.3& 28.5& 60.4& 84.6& 58.5& 17.0& 15.8& 48.0& 91.8& 58.1& 21.6& 16.1& 56.4\\
Mb2C \cite{3681292}& MM'2024& 95.1& 73.0& 60.6& 28.1& 60.1& 85.9& 62.3& 19.7& 19.2& 44.7& 92.5& 62.7& 24.2& 20.6& 55.9\\
NICE \cite{song2024decodingnaturalimageseeg}&ICLR'2024&96.3& 82.9& 66.0& 27.1& 59.2& 84.3& 66.5& 22.1& 21.6& 46.8& 92.4& 66.8& 27.0& 22.2& 57.8\\
ATM-S \cite{NEURIPS2024_ba5f1233}&NIPS'2024& 96.8& 80.4& 64.2& 25.5& 58.0& 84.5& 70.8& 24.7& 18.7& 43.1& 92.9& 71.0& 29.9& 25.6& 51.9\\
UBP \cite{11094846}& CVPR'2025& 97.6& 77.2& 61.4& 33.0& 59.5& 85.9& 72.5& 28.0& 26.7& 55.2& 93.9& 75.5& 33.1& 28.7& 59.3\\
CogCap \cite{zhang2024cognitioncapturerdecodingvisualstimuli}& AAAI'2025& 97.3& 78.0& 62.1& 37.5& 65.5& 86.3& 77.2& 31.6& 22.7& 48.9& 95.3& 79.9& 36.4& 32.6& 62.5\\
ViEEG \cite{liu2025vieeghierarchicalvisualneural}& MM'2025& 97.9& 90.1& 66.9& 40.5& 72.0& 87.9&81.5& 34.8& 25.0& 54.0& 96.6& 84.7& 40.0& 35.8& 78.0\\
FLORA \cite{li2025brainflorauncoveringbrainconcept}& MM'2025& 97.6& 88.7& 65.7& 40.9& 74.5& 89.3& 83.4& 38.3& 24.5& 52.6& 97.8& 89.2& 44.0& 38.0& 76.5\\
\hline
Bratrix (Ours)& /&98.4& 97.9& 70.9& 51.5& 84.5& 90.7& 89.0& 43.4& 26.8& 56.0& 98.6& 92.8&48.8& 44.5& 80.6\\ 
\rowcolor{blue!10} Bratrix-M (Ours) &/& \textbf{99.0} & \textbf{98.6} & \textbf{72.3} & \textbf{55.3} & \textbf{86.9} & \textbf{91.6} & \textbf{90.3} & \textbf{44.9} & \textbf{27.3} & \textbf{58.2} & \textbf{99.2} & \textbf{93.6} & \textbf{49.9} & \textbf{46.1} & \textbf{81.3} \\
\hline
\multicolumn{16}{c}{\textbf{Subject-independent}} \\ 
\hline
LSTM \cite{vennerød2021longshorttermmemoryrnn}&/& 57.0& 42.6& 11.3& 7.9& 16.1& 39.4 & 29.6& 2.0& 1.8& 5.3& 38.4& 12.9& 1.8& 1.9& 5.8\\
ConvNet \cite{radford2021learningtransferablevisualmodels}&/& 61.5& 44.9& 14.3& 8.2& 16.9& 41.4& 31.9& 2.2& 2.0& 5.9& 40.2& 13.6& 2.0& 2.0& 6.0\\
EEGNet \cite{Lawhern_2018}&/& 64.5& 45.7& 18.1& 9.0& 17.6& 43.4& 32.3& 2.4& 2.2&6.1& 42.0& 14.2& 2.2& 2.1& 6.4\\
\midrule
MindEyev2 \cite{scotti2024mindeye2sharedsubjectmodelsenable}& ICML'2024& 78.6& 46.9& 21.1& 6.1& 13.0& 42.2& 33.6& 5.9& 1.8& 4.1& 39.9&15.3& 4.5& 2.1& 4.6\\
BraVL \cite{du2023decodingvisualneuralrepresentations}& TPAMI'2023& 83.8& 47.2& 24.9& 6.0& 13.1& 43.7& 32.9& 6.3& 2.0& 4.3& 41.2& 15.9& 4.8& 2.2& 4.6\\
Mb2C \cite{3681292}& MM'2024& 78.4& 48.9& 21.2& 17.2& 32.3& 48.1& 33.8& 7.1& 3.3& 6.7& 44.8& 16.3& 5.4& 3.4& 6.9\\
NICE \cite{song2024decodingnaturalimageseeg}& ICLR'2024& 81.9& 50.2& 22.3& 19.9& 36.3& 51.6& 34.8& 7.9& 3.8& 7.5& 49.1& 17.3& 6.1& 4.0& 7.6\\
ATM-S \cite{NEURIPS2024_ba5f1233}& NIPS'2024& 78.9& 52.7& 20.2& 17.9& 34.4& 55.4& 37.7& 8.6& 3.4& 7.0& 52.8& 19.1& 6.7& 3.7& 7.1\\
UBP \cite{11094846}& CVPR'2025& 76.4& 58.2& 17.2& 16.4& 29.9& 59.9& 38.2& 9.3& 2.2& 10.4& 56.9& 21.0& 7.3& 3.6& 7.2\\
CogCap \cite{zhang2024cognitioncapturerdecodingvisualstimuli}& AAAI'2025& 77.3& 68.7&18.1& 16.3& 30.3& 64.7& 42.3& 9.9& 4.7& 9.8& 61.5& 24.6& 7.9& 5.0& 10.1\\
ViEEG \cite{liu2025vieeghierarchicalvisualneural}& MM'2025& 88.7& 70.9& 25.8& 19.0& 41.5& 68.7& 49.0& 10.3& 3.5& 8.6& 65.2& 27.3& 8.1& 3.7& 9.0\\
FLORA \cite{li2025brainflorauncoveringbrainconcept}& MM'2025& 87.0& 72.3& 24.1& 18.9& 37.9& 71.9& 52.6& 10.4& 3.4& 8.2& 68.3& 29.6& 8.1& 3.6& 8.7\\
\hline
Bratrix (Ours)&/& 91.8& 80.9& 32.1& 20.5& 45.3& 80.3& 56.6& 11.5& 4.9& 14.5& 70.1& 32.2& 8.9& 4.2& 12.3\\
\rowcolor{blue!10} Bratrix-M (Ours)& /&\textbf{93.0} & \textbf{83.1} & \textbf{34.5} & \textbf{23.5} & \textbf{47.9} & \textbf{83.2} & \textbf{58.9} & \textbf{14.7} & \textbf{5.6} & \textbf{17.2} & \textbf{72.3} & \textbf{34.0} & \textbf{10.1} & \textbf{4.5} & \textbf{15.1} \\
\hline
\end{tabular}} 
\label{tab:comparison}
\end{table*}

\begin{table*}[t]
\centering
\scriptsize
\caption{Ablation Performance of Bratrix on THINGS-EEG2, THINGS-MEG, and THINGS-fMRI datasets under subject-dependent settings. Ablation experiments include language semantic decoupling ablation, vision semantic decoupling ablation, and module ablation.}
\renewcommand{\arraystretch}{1} 
\setlength{\tabcolsep}{3pt}
\resizebox{\textwidth}{!}{%
\begin{tabular}{lccccc|ccccc|ccccc}
\hline
\multirow{3}{*}{Model} 
& \multicolumn{5}{c}{\textbf{THINGS-EEG2}} 
& \multicolumn{5}{c}{\textbf{THINGS-MEG}} 
& \multicolumn{5}{c}{\textbf{THINGS-fMRI}} \\
\cline{2-16}
&\shortstack{2-way \\ Top-1} & \shortstack{4-way \\ Top-1} & \shortstack{50-way \\ Top-1} & \shortstack{200-way \\ Top-1} & \shortstack{200-way \\ Top-5}
& \shortstack{2-way \\ Top-1} & \shortstack{4-way \\ Top-1} & \shortstack{10-way \\ Top-1} & \shortstack{200-way \\ Top-1} & \shortstack{200-way \\ Top-5}
& \shortstack{2-way \\ Top-1} & \shortstack{4-way \\ Top-1} & \shortstack{10-way \\ Top-1} & \shortstack{100-way \\ Top-1} & \shortstack{100-way \\ Top-5} \\
\hline
\multicolumn{16}{c}{\textbf{Language Semanctic Ablation Performance}} \\ 
\hline
1. w. Primary & 97.30 & 88.30 & 67.90 & 45.10 & 76.70 & 90.50 & 82.90 & 38.70 & 26.20 & 49.40 & 98.30 & 87.10 & 41.80 & 39.00 & 79.60 \\
2. w. Concrete & 98.20 & 93.60 & 68.80 & 48.60 & 80.90 & 90.60 & 85.80 & 41.50 & 26.40 & 52.50 & 98.40 & 88.80 & 45.20 & 42.10 & 80.20 \\
3. w. Psychic& 98.30 & 96.70 & 69.30 & 50.10 & 84.10 & 90.70 & 88.30 & 42.70 & 26.60 & 54.70 & 98.60 & 92.10 & 47.90 & 44.10 & 80.50 \\
4. w. Physical & 98.40 & 97.90 & 70.90 & 51.50 & 84.50 & 90.70 & 89.00 & 43.40 & 26.80 & 56.00 & 98.60 & 92.80 & 48.80 & 44.50 & 80.60 \\
\hline
\multicolumn{16}{c}{\textbf{Visiual Semanctic Ablation Performance}} \\ 
\hline
1. w. Background& 97.20 & 87.90 & 65.20 & 42.90 & 73.40 & 87.70 & 78.00 & 36.20 & 24.40 & 52.40 & 96.30 & 83.50 & 39.50 & 36.40 & 77.20 \\
2. w. Foreground & 97.50 & 90.60 & 67.90 & 45.90 & 77.30 & 89.20 & 82.10 & 37.70 & 25.30 & 53.00 & 97.00 & 87.30 & 43.5 & 40.30 & 78.50 \\
3. w. Depth& 97.80 & 95.10 & 68.80 & 49.40 & 82.90 & 90.00 & 85.60 & 41.10 & 26.00 & 54.50 & 98.30 & 90.50 & 44.10 & 42.20 & 80.10 \\
4. w. Color & 98.20 & 96.40 & 69.30 & 50.20 & 83.90 & 90.50 & 87.80 & 42.30 & 26.50 & 55.40 & 98.40 & 92.10 & 47.10 & 43.40 & 80.50 \\
5. w. Texture & 98.40 & 97.90 & 70.90 & 51.50 & 84.50 & 90.70 & 89.00 & 43.40 & 26.80 & 56.00 & 98.60 & 92.80 & 48.80 & 44.70 & 80.60 \\
\hline
\multicolumn{16}{c}{\textbf{Module Ablation Performance}} \\ 
\hline
Baseline & 97.20 & 92.30 & 66.80 & 44.90 & 79.40 & 90.50 & 87.60 & 39.40 & 23.90 & 53.90 & 98.40 & 87.60 & 42.90 & 39.40 & 78.90 \\
1. w. SUP & 97.90 & 94.60 & 68.30 & 46.90 & 82.60 & 90.60 & 88.30 & 41.60 & 24.60 & 54.60 & 98.40 & 88.40 & 44.10 & 40.60 & 79.60 \\
2. w. SLA  & 98.30 & 97.60 & 69.40 & 50.40 & 83.90 & 90.70 & 88.90 & 43.00 & 26.30 & 55.80 & 98.50 & 92.60 & 47.60 & 43.50 & 80.40 \\
3. w. $\mathcal{L}_{kl}$ (Bratrix) & 98.40 & 97.90 & 70.90 & 51.50 & 84.50 & 90.70 & 89.00 & 43.40 & 26.80 & 56.00 & 98.60 & 92.80 & 48.80 & 44.40 & 80.60 \\
\hline
\hline
\end{tabular}}
\label{tab:ablation}
\end{table*}
\section{Experiments and Results}
\label{sec:results}
\begin{figure*}[t]
    \centering
    \includegraphics[width=1\linewidth]{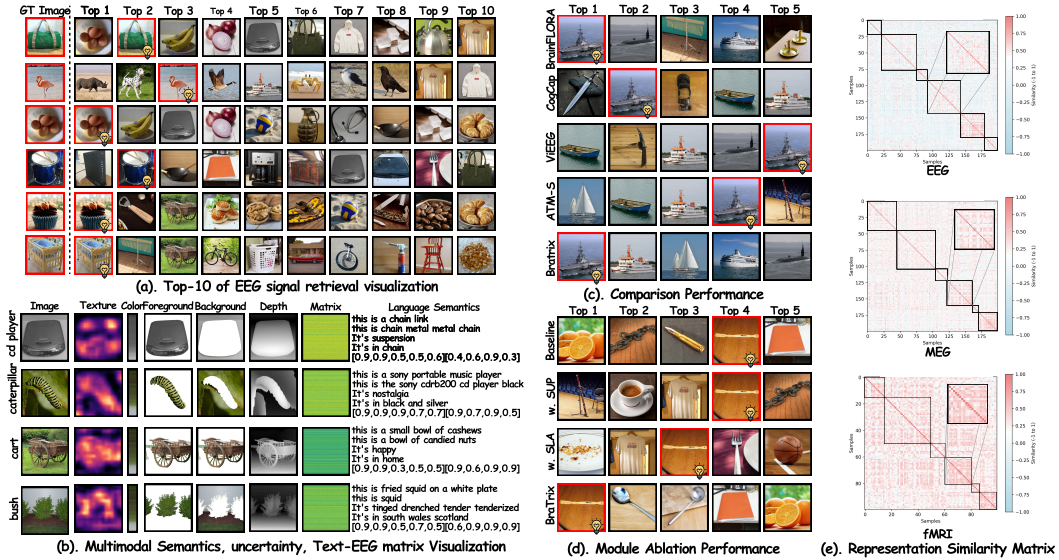}
    \caption{(a). Zero-shot Top-10 Performance of EEG signal retrieval visualization. (b). Example of multimodal semantics, uncertainty, and semantic matrix visualization. (c). Zero-shot Top-5 Performance of comparable methods. (d). Zero-shot Top-5 Performance of ablation experiment. (e).  Representational similarity matrices (RSM) of brain neural signals across categories (Tool, Food, Clothes, Vehicle, Animal, and Others), and zoomed-in view of Food category.}
    \label{fig:3}
\end{figure*}
\subsection{Dataset and Implementation Details}
\textbf{THINGS-EEG}~\cite{GIFFORD2022119754} includes recordings from 10 subjects under a Rapid Serial Visual Presentation paradigm. The training set contains 1,654 concepts (10 images each, four repetitions), while the test set comprises 200 unseen concepts (one image, 80 repetitions). EEG was recorded with a 64-channel cap at 1,000~Hz, band-pass filtered (0.1-100~Hz), downsampled to 100-250~Hz, segmented (-200 to 800~ms), baseline-corrected, and repetition-averaged to enhance signal-to-noise ratio.\\
\textbf{THINGS-MEG}~\cite{82580} includes recordings from four subjects using 271 channels. The training set comprises 1,854 concepts (12 images each, viewed once), and the test set contains 200 unseen concepts (one image, 12 repetitions). Standard MEG preprocessing was applied, with trial averaging to improve signal-to-noise ratio.\\
\textbf{THINGS-fMRI}~\cite{82580} provides data from three subjects. The training set includes 720 concepts (12 images each, viewed once), while the test set consists of 100 unseen concepts (one image, 12 repetitions). fMRI volumes were collected with a 7T scanner and preprocessed using standard pipelines with trial averaging to reduce noise.\\
\textbf{Implementation Details. }All experiments are implemented on PyTorch with one NVIDIA RTX4090 GPU. For vision, Brain, and Language Encoder, we employed CLIP (ViT-L-14) \cite{radford2021learningtransferablevisualmodels}, iTransformer \cite{liu2024itransformerinvertedtransformerseffective}, and BLIP \cite{li2022blipbootstrappinglanguageimagepretraining} to generate image and brain embeddings, respectively. Each modality encoder was trained on the original training set of the THINGS dataset for 60 epochs in unimodal setting and 30 epochs in multi-modal setting, with a learning rate of 1e-4 and 5e-5. The batch size is 250, utilizing the AdamW optimizer. The feature alignment fine-tuning of image generation and captioning tasks is conducted based on the Stable Diffusion XL \cite{podell2023sdxlimprovinglatentdiffusion} and shikra-7B \cite{chen2023shikraunleashingmultimodalllms} with 150 epochs and 240 epochs, respectively. The learning rate adjustment strategy starts with a linear warm-up for 5\% of the total steps, followed by cosine annealing. The metrics for reconstruction and caption tasks followed BrainHub \cite{5_14}. The visual semantics followed Depth Anything \cite{yang2024depthv2}, BiRefNet \cite{Zheng_2024}, and Resnet \cite{he2015deepresiduallearningimage} (see more details in Appendix).
\section{Results}
\subsection{Image Retrieval Results}
\textbf{Comparison Performance: }We conduct comparison experiments on three datasets under both subject-dependent and subject-independent settings. As shown in Table \ref{tab:comparison}, Bratrix achieves state-of-the-art performance across 2-way, 4-way, 50-way, and 200-way classification tasks. Taking the 200-way task as the example, our method surpasses the latest methods by 16.6\%, 10.64\%, and 4.2\% on the three datasets under the subject-dependent setting. Qualitative results on the EEG modality, presented in Fig. \ref{fig:3} (a), demonstrate that Bratrix establishes rich semantic associations and retrieves top-10 samples with high semantic relevance, which can be attributed to the hierarchical perception of deep language–visual semantics illustrated in Fig. \ref{fig:3} (b). Moreover, as shown in Fig. \ref{fig:3} (c), Bratrix outperforms competing methods in the top-5 retrieved samples, all of which correspond to the "ship" category. More retrieval examples in the other two modalities (MEG and fMRI) is shown in Fig. \ref{fig:6} (a). Additional qualitative and quantitative comparisons are provided in Appendix and \ref{app:Qualitative_Retrieval}. Overall, these extensive results demonstrate that Bratrix consistently captures fine-grained semantic structure and achieves superior brain–vision alignment across modalities and tasks.

\begin{figure*}[t]
    \centering
    \includegraphics[width=1\linewidth]{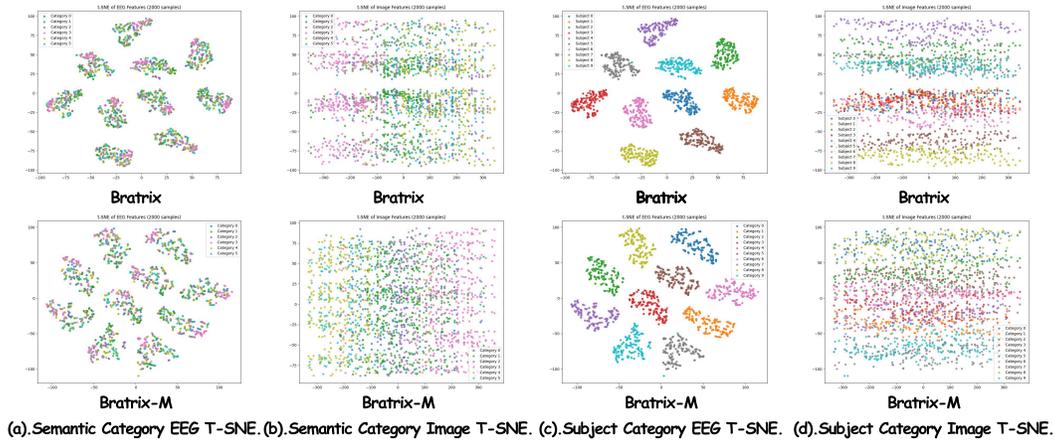}
    \caption{T-SNE Visualization \cite{vandermaaten08a} of Bratrix and Bratrix-M with semantic categories and subject categories.}
    \label{fig:4}
\end{figure*}
\begin{table}[t]
\centering
\caption{Quantitative multi-modal Visual Reconstruction results.}
\begin{tabular}{lcccc}
\toprule
Modality & Inception$\uparrow$ & CLIP$\uparrow$ & EffNet-B$\downarrow$ & SwAV$\downarrow$ \\
\midrule
EEG & 0.591 & 0.659 & 0.886 & 0.655\\
MEG  & 0.631 & 0.638 & 0.771 & 0.643 \\
fMRI & 0.654 & 0.667 & 0.769 & 0.627 \\
\bottomrule
\end{tabular}
\label{tab:3}
\caption{Quantitative multi-modal Visual Caption results.}
\begin{tabular}{lcccc}
\toprule
Modality  & BLEU4$\uparrow$ & METEOR$\uparrow$ & ROUGE$\uparrow$ & CIDEr$\uparrow$ \\
\midrule
EEG  & 22.02 & 23.48 & 48.96 & 3.26 \\
MEG  & 25.34 & 24.71 & 51.61 & 3.87 \\
fMRI & 24.28 & 25.84 & 50.07 & 4.02 \\
\bottomrule
\end{tabular}
\label{tab:4}
\end{table}
\textbf{Ablation Performance: }We further conduct ablation experiments on three datasets under the subject-dependent setting. As shown in Table \ref{tab:ablation}, we evaluate three levels of ablation: (i) language semantic decoupling, (ii) vision semantic decoupling, and (iii) module-level ablation. Results indicate that the incremental integration of Bratrix modules consistently improves performance across 2-way, 4-way, 50-way, and 200-way retrieval tasks, highlighting both the rationality of the overall architecture and the effectiveness of semantic decomposition in the language and vision modalities. Notably, the integration of language semantics yields relatively stable performance gains, whereas the integration of visual semantics produces less stable overall positive improvements. In module ablation, the combination of the SUP and SLA modules leads to significant improvement, while the KL-divergence loss provides further refinement in alignment precision. We also conduct qualitative ablation visualizations. As shown in Fig. \ref{fig:3} (d), retrieval results progressively improve from top-4 to top-1 predictions, and semantic relevance among the top-5 samples increases with the integration of additional modules, demonstrating the effectiveness of the proposed components. More ablation visualizations are provided in Appendix.
\begin{figure}[t]
  \centering
  \begin{minipage}[t]{0.47\linewidth}
    \centering
    \includegraphics[width=\linewidth]{1000006211.pdf}
    \caption{(a) Zero-shot comparison across 10 subjects and (b) comparison between coarse and aligned EEG.}
    \label{fig:5}
  \end{minipage}
  \hfill
  \begin{minipage}[t]{0.48\linewidth}
    \centering
    \includegraphics[width=\linewidth]{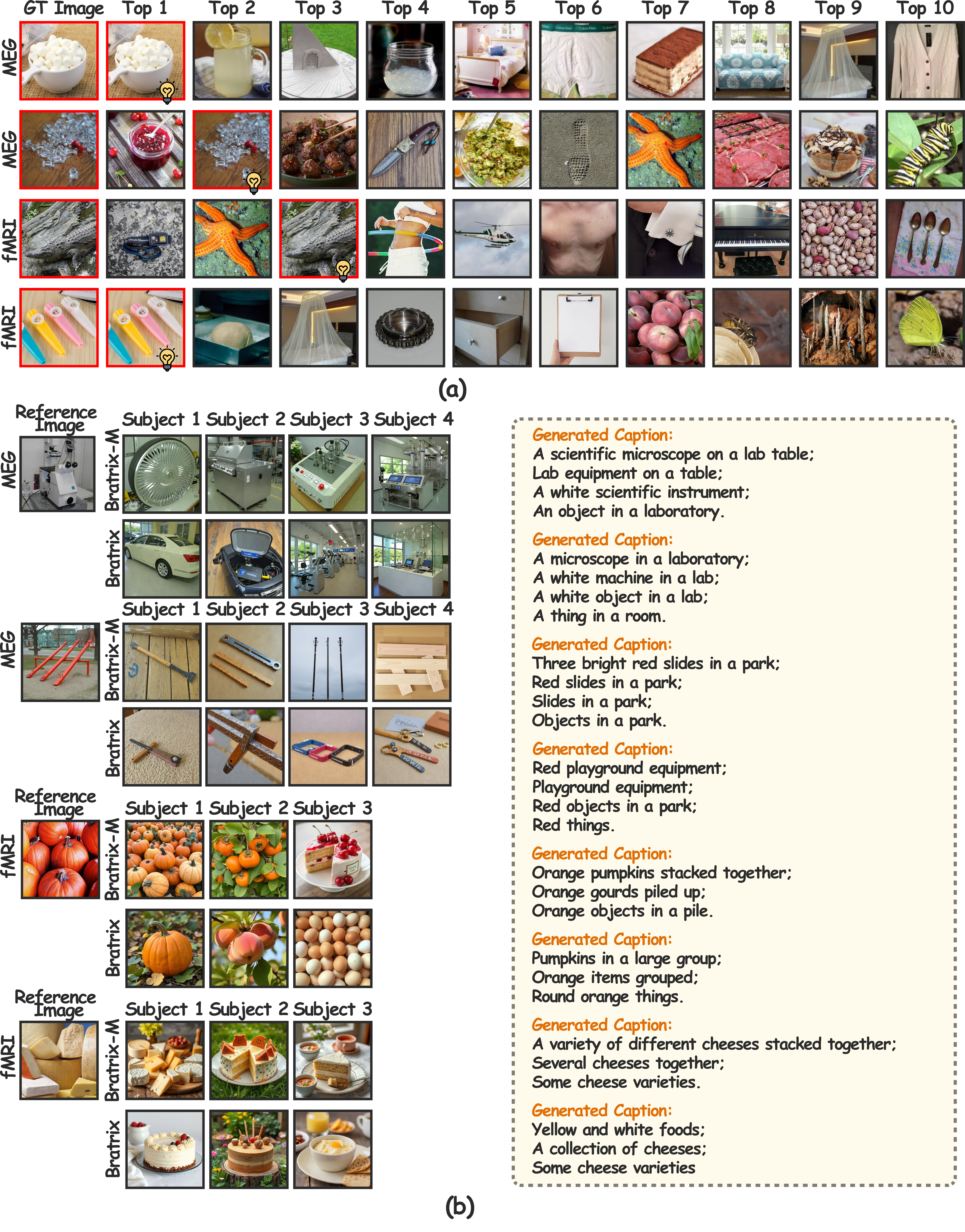}
    \caption{(a) Zero-shot retrieval Top-10 performance in MEG/fMRI and (b) comparison between Bratrix and Bratrix-M.}
    \label{fig:6}
  \end{minipage}
\end{figure}

\textbf{Visualization Results: }Finally, we conduct visualization experiments to analyze the semantic alignment performance of Bratrix. As shown in Fig.~\ref{fig:3} (e), we divide 200 test images into six initial categories (Tool, Food, Clothes, Vehicle, Animal, and Others), and visualize their representational similarity matrices (RSMs) in the EEG modality. Clear differences in category-level weights can be observed, indicating that Bratrix is able to capture stable coarse-grained semantic distinctions. Also, in modalities such as MEG and fMRI, inter-category similarities remain relatively high. This observation suggests that early-stage semantic categorizations can be ambiguous, which may be attributed to the influence of language anchors, resembling the way humans may experience similar instantaneous responses, such as fear or tension, to perceptually distinct stimuli like warships and bullets. More image RSMs is shown in Appendix. Beyond RSMs, we visualize the distribution of t-SNE across 10 subjects in the subject-independent setting. As shown in Fig.~\ref{fig:4} (a), Bratrix organizes the semantic representations of each subject into coherent subject-specific clusters, whereas Bratrix-M produces a more dispersed embedding distribution, demonstrating enhanced cross-subject unification of EEG semantic spaces. In the semantic-category setting, Bratrix yields relatively consistent semantic spaces for visual representations, while Bratrix-M further strengthens this consistency, as shown in Fig.~\ref{fig:4} (b). This result highlights its superior capability in aligning visual and neural representations. Finally, under the subject-category setting, we visualize both EEG signals and image embeddings. As shown in Fig.~\ref{fig:4} (c) and Fig.~\ref{fig:4} (d), Bratrix-M produces more unified and dispersed representations across different participants for both modalities compared to Bratrix, demonstrating that the proposed language-anchored semantic matrices generalize effectively to cross-modal alignment tasks and reinforce semantic space unification.
\subsection{Image Reconstruction and Caption Results}
We further evaluate Bratrix on image reconstruction and captioning tasks. As shown in Table~\ref{tab:3} and Table~\ref{tab:4}, Bratrix achieves consistently strong results across all metrics. Notably, EEG performance is relatively weaker compared to MEG and fMRI, which may reflect that fine-grained semantic representations reduce sensitivity to coarse category-level neural patterns. To gain deeper insights, we visualize reconstruction and captioning results across modalities. Figure~\ref{fig:5} (a) highlights substantial inter-subject variability in EEG performance. In Figure~\ref{fig:5} (b), we compare reconstructions from direct image–EEG alignment with those from Bratrix’s language-anchored alignment. The latter produces substantially refined reconstructions, demonstrating the benefit of language-guided semantic alignment. Figure~\ref{fig:5} (c) further shows captioning results in the EEG modality, while Figure~\ref{fig:6} (b) illustrates reconstruction and captioning outcomes in MEG and fMRI, where performance surpasses EEG and aligns with quantitative results, albeit with significant subject-level variability. Finally, comparison between Bratrix and Bratrix-M in Figure~\ref{fig:6} (b) demonstrates that multimodal fine-tuning strategy shows more robust and stable performance in both reconstruction and captioning.
\section{Conclusion, Limitation, and Future Work}
In this work, we introduced \textbf{Bratrix}, a language-anchored brain–visual semantic alignment framework. Through semantic decoupling modules and an uncertainty perception mechanism for visual and language semantics, Bratrix constructs language-anchored semantic matrices that enable fine-grained cross-modal correspondence between neural activity and perceptual representations. A two-stage training strategy improves multi-modal performance, yielding the extended Bratrix-M. Extensive experiments on EEG, MEG, and fMRI benchmarks show that Bratrix consistently outperforms state-of-the-art methods in retrieval, reconstruction, and caption tasks, while ablation and visualization studies validate its core components. Despite these advances, several limitations remain. First, inter-subject variability remains a core challenge, motivating future work on subject-invariant modeling. Second, our experiments focus on static visual stimuli, while real-world cognition involves dynamic and multimodal experiences; extending Bratrix to spatiotemporal and interactive scenarios like Vision-Language-Action model is a promising direction. Finally, integrating Bratrix into closed-loop brain–computer interfaces could enable practical applications in assistive technologies and cognitive neuroscience. We leave these avenues for future exploration.

\bibliography{iclr2026_conference}
\bibliographystyle{iclr2026_conference}
\clearpage
\appendix
\begin{figure*}[h]
    \centering
    \includegraphics[width=0.8\linewidth]{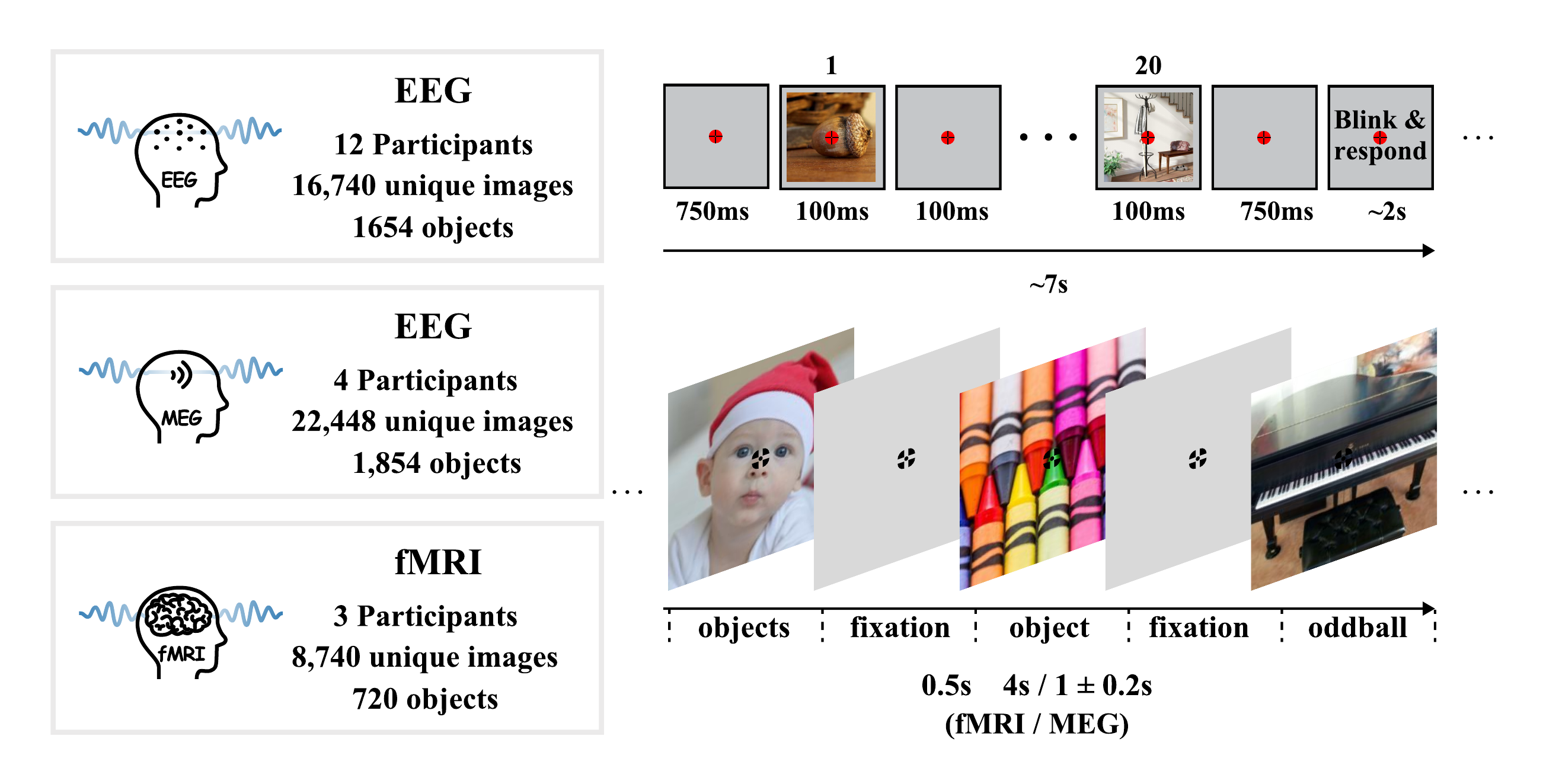}
    \caption{Dataset set up of THINGS-EEG, THINGS-MEG, and THINGS-fMRI dataset.}
    \label{fig:dataset}
\end{figure*}
\section{Details of the Datasets}
\label{app:Dataset_Details}
As shown in Fig. \ref{fig:dataset}, the THINGS dataset is a large-scale, multi-modal benchmark designed to study how the human brain and artificial models represent visual objects. It includes three neural recording modalities (EEG, MEG, and fMRI) collected from 19 participants who viewed a wide range of natural object images spanning 1,854 categories and over 26,000 exemplars. The dataset is divided into training and test splits for modeling and evaluation. The zero-shot test set is provided to assess the model’s capacity to generalize to unseen object categories without direct supervision.
\subsection{THINGS-EEG Dataset}
THINGS-EEG is a large-scale, high-temporal-resolution electroencephalography (EEG) dataset focused on human visual object recognition, with multiple versions and subsets of distinct specifications. The core version comprises 50 subjects, while subsets such as THINGS-EEG2 consist of 10 healthy adults (mean age 28.5 years, 8 females, 2 males). All stimuli are derived from the THINGS database: the core version covers 1854 object concepts and 22,248 natural-background images, THINGS-EEG2 contains 16,740 images, and each object concept is matched with no fewer than 12 images to ensure stimulus diversity. In terms of trials and samples, each participant in the 10-subject subset completed 82,160 trials (16,540 training image conditions with 4 repetitions each, 200 test image conditions with 80 repetitions each), resulting in a total of 160,000 samples (16,000 per participant).
The preprocessing workflow follows a standardized procedure: stimuli are presented via the Rapid Serial Visual Presentation (RSVP) paradigm (100 ms per image, 100 ms blank interval, pseudorandom order) to eliminate anticipatory interference. For signal processing, 17 posterior channels (occipital/parietal lobes, visual-function relevant) are selected from 63 original acquisition channels to filter irrelevant interference; EEG signals within the -200 ms to 800 ms recording window are downsampled to 100 Hz or 250 Hz to reduce redundancy. After excluding target stimulus trials, multivariate noise normalization is independently applied to each session’s data, followed by baseline correction and standardization to optimize signal quality. Finally, structured data matrices are generated (training set: 16540×4×17×100, i.e., image conditions × repetitions × channels × time points; test set: 200×80×17×100) with comprehensive metadata (image concepts, repetition IDs, participant information).
\subsection{THINGS-MEG Dataset}
The THINGS-MEG dataset comprises 4 healthy volunteers (2 females, 2 males) with a mean age of 23.25 years at the study onset, all of whom completed 12 main MEG sessions. Stimuli are drawn from the THINGS database, encompassing 1,854 object concepts (each paired with 12 unique images) and totaling 22,448 distinct object images, along with 200 repeatedly presented test images and 100 distractor images. Each image is displayed for 500 ms, followed by a variable blank interval of 1,000±200 ms; participants are instructed to maintain central fixation and perform an orthogonal oddball detection task (responding to distractor images) to ensure sustained attention. Data acquisition employs a CTF 275 MEG system, which includes a whole-head array of 275 radial 1st-order gradiometers/SQUIDs. Three dysfunctional channels (MLF25, MRF43, MRO13) are excluded, resulting in 272 valid channels. The sampling rate is set to 1,200 Hz, and online 3rd gradient balancing is applied to reduce background noise. Eye-tracking data (gaze position and pupil size) are simultaneously recorded at 1,200 Hz and stored in miscellaneous MEG channels; an optical sensor is used to account for temporal delays between stimulus display and data acquisition. Preprocessing is conducted using MNE-Python: raw data for each run undergo band-pass filtering within the 0.1–40 Hz range; transient signal dropouts (duration $<$200 ms) in Participants M1 and M2 are replaced with the median sensor response; trials are epoched from –100 ms to 1,300 ms relative to stimulus onset (marked by parallel port triggers and optical sensor signals); baseline correction is applied by subtracting the mean and normalizing by the standard deviation of the baseline period (–100 ms to stimulus onset); one consistently noisy channel (MRO11) is excluded for all participants; and data are downsampled to 200 Hz to reduce computational load. To minimize head motion, participants wear custom head casts—median within-session head motion is $<$1.5 mm, and median inter-session head motion is $<$3 mm.
\subsection{THINGS-fMRI Dataset}
The THINGS-fMRI dataset includes 3 healthy volunteers (2 females, 1 male) with a mean age of 25.33 years at study onset, all completing 12 main functional sessions. Stimuli are from the THINGS database, covering 720 object concepts (12 unique images per concept) for 8,740 distinct object images, plus 100 repeatedly presented test images and 100 BigGAN-generated distractor images.
Data acquisition uses a 3 Tesla Siemens Magnetom Prisma scanner with a 32-channel head coil. Key functional parameters are 2 mm isotropic resolution, repetition time (TR) 1.5 s, echo time (TE) 33 ms, 75° flip angle, and 3× multi-band slice acceleration (posterior-to-anterior phase encoding). Complementary data (high-resolution T1/T2-weighted anatomical images, vascular data, resting-state functional data) are also acquired.
Preprocessing is done via fMRIPrep (v20.2.0) following a standardized pipeline: slice timing correction, rigid-body head motion correction, susceptibility distortion correction (gradient-echo field maps), spatial alignment to T1-weighted images, and brain tissue segmentation. A custom ICA-based denoising approach enhances quality—functional runs undergo smoothing (FWHM=4 mm) and high-pass filtering (120 s cut-off) before ICA; components are labeled as signal/noise by two raters, with univariate thresholds classifying all 20,388 components. Finally, a single-trial regularized GLM estimates BOLD amplitudes: runs are converted to percent signal change, residualized against regressors; optimal HRFs (from 20 options) are selected per voxel, fractional ridge regression mitigates overfitting, and beta coefficients are rescaled for final single-trial responses.
\section{Details of Deep Semantic Extraction}
\subsection{Details of Deep Visual Semantics}
Visual semantics are extracted across five complementary levels: foreground, background, depth, color, and texture. Specifically:

\begin{itemize}
    \item \textbf{Foreground and background}: The original RGB images are masked using a high-resolution dichotomous segmentation model~\cite{Zheng_2024} to obtain foreground and background regions. Each masked image is resized to $224 \times 224$ and encoded via a pretrained visual-language model, producing normalized embeddings that capture semantically meaningful features for both regions.
    
    \item \textbf{Depth}: Depth maps are obtained from a state-of-the-art depth estimation model~\cite{yang2024depthv2}. The three-channel depth images are converted to grayscale using standard luminance weighting, down-sampled via adaptive average pooling to $32 \times 32$, flattened, and L2-normalized to generate compact depth embeddings.
    
    \item \textbf{Color}: The original RGB images are down-sampled to $32 \times 32$ and flattened across channels. Pixel values are summed across the R, G, and B channels, and the resulting vector is normalized, producing a compact color feature embedding.
    
    \item \textbf{Texture}: Texture information is derived from the fourth residual block of ResNet~\cite{he2015deepresiduallearningimage}. The spatial feature maps are averaged across the spatial dimensions, split into two halves along the channel dimension, summed, and normalized to produce a 1024-dimensional texture embedding.
    
    \item \textbf{Integration}: The embeddings from all five semantic levels—foreground, background, depth, color, and texture—are stacked into a unified feature tensor of shape $[B, 5, 1024]$, providing a multi-level, normalized representation of visual content for subsequent multimodal processing.
\end{itemize}
\subsection{Details of Deep Language Semantics}
Deep Language Semantics are extracted across four complementary levels: detail, psychic, coarse, and locational. They are inferred from a state-of-the-art caption model BLIP~\cite{li2022blipbootstrappinglanguageimagepretraining}, which is designed to learn joint representations of images and text for tasks, such as image captioning and visual question answering:
\begin{itemize}
    \item \textbf{Detail-level Semantics}: Captures the explicit, fine-grained content of the image. Generated using the prompt: 
    \texttt{this is a picture of} \\
    The model produces a concise textual description of the visual objects or elements present in the image, which is then cleaned, lowercased, and stripped of punctuation to form a normalized representation.

    \item \textbf{Psychic-level Semantics}: Captures the emotional or mental impression elicited by the image. Generated using the prompt:
    \texttt{Question: Describe the mental feeling of this picture. Answer:} \\
    This semantic representation encodes the affective response to the visual content, allowing the model to reason about the psychological impact or mood conveyed by the image.

    \item \textbf{Coarse-level Semantics}: Captures a high-level, abstract characterization of the image content. Generated using the prompt:
    \texttt{Question: Describe this picture in three words and only words. Answer:} \\
    The output is constrained to three words, providing a succinct, abstract representation that summarizes the image's main concepts or themes.

    \item \textbf{Locational Semantics}: Captures spatial or situational context of the image. Generated using the prompt:
    \texttt{Question: Describe the location of this picture. Answer:} \\
    This semantic embedding encodes information about the environment, setting, or geographic context of the scene, complementing other semantic levels.
\end{itemize}
\section{Compared Methods Introduction}
\begin{itemize}
\item \textbf{LSTM \cite{vennerød2021longshorttermmemoryrnn}: }a recurrent neural network architecture designed to capture long-range temporal dependencies by leveraging gated mechanisms that regulate information flow.\\
\item \textbf{ConvNet \cite{radford2021learningtransferablevisualmodels}: }a large-scale vision–language model that jointly learns image and text representations through contrastive pretraining, enabling robust zero-shot transfer across diverse visual recognition tasks.
\item \textbf{EEGNet \cite{Lawhern_2018}: }a compact convolutional neural network for EEG-based BCI that employs depthwise separable convolutions to reduce overfitting and enhance generalization.\\
\item \textbf{MindEye2 \cite{scotti2024mindeye2sharedsubjectmodelsenable}: }a subject-generalizable brain-to-image reconstruction framework that aligns fMRI signals into a shared latent space and maps them to pixel space via CLIP-guided Stable Diffusion, enabling high-quality reconstructions. \\
\textbf{BraVL \cite{du2023decodingvisualneuralrepresentations}: }a multimodal brain–vision–language decoding framework that leverages mixture-of-product-of-experts generative modeling and mutual information maximization to enable data-efficient cross-modal alignment and novel-category decoding.\\
\item \textbf{MB2C \cite{3681292}: }a multimodal brain–vision decoding framework that employs dual-GAN bidirectional cycle consistency to align EEG and visual embeddings in a shared semantic space, enabling robust zero-shot decoding and state-of-the-art performance on classification and reconstruction tasks.\\
\item \textbf{NICE \cite{song2024decodingnaturalimageseeg}: }a self-supervised EEG-to-image framework that uses contrastive learning and attention modules to align EEG and visual features, enabling zero-shot object recognition and interpretable neural decoding.\\
\item \textbf{ATM-S \cite{NEURIPS2024_ba5f1233}: }an end-to-end EEG-based visual reconstruction framework that projects neural signals into a shared CLIP embedding space and employs a two-stage diffusion pipeline for zero-shot image generation, achieving state-of-the-art decoding, retrieval, and reconstruction.\\
\item \textbf{UBP \cite{11094846}: }an uncertainty-aware brain-to-image decoding framework that dynamically blurs high-frequency image details based on estimated signal–stimulus mismatch, improving zero-shot retrieval and alignment under limited paired data.\\
\item \textbf{CogCap \cite{zhang2024cognitioncapturerdecodingvisualstimuli}: }a unified EEG decoding framework that leverages modality-specific expert encoders and a diffusion prior to map EEG embeddings into CLIP space, enabling high-fidelity visual reconstruction without generative model fine-tuning.\\
\item \textbf{ViEEG \cite{liu2025vieeghierarchicalvisualneural}: }a neuro-inspired EEG decoding framework that models hierarchical visual processing via three-stream EEG encoders and cross-attention routing, aligned with EEG-CLIP representations for zero-shot object recognition.\\
\item \textbf{FLORA \cite{li2025brainflorauncoveringbrainconcept}: }a unified framework for integrating EEG, MEG, and fMRI using multimodal large language models with modality-specific adapters, constructing shared neural representations for cross-modal visual retrieval and cognitive analysis.\\
\end{itemize}
These methods represent a diverse set of strategies for EEG decoding and visual reconstruction, serving as baselines for benchmarking Bratrix. Notably, Bratrix introduces language semantics into the alignment process and achieves superior performance in a multimodal setting, demonstrating the complementary benefits of leveraging shared semantic spaces across modalities.
\begin{figure*}[t]
    \centering
    \includegraphics[width=1\linewidth]{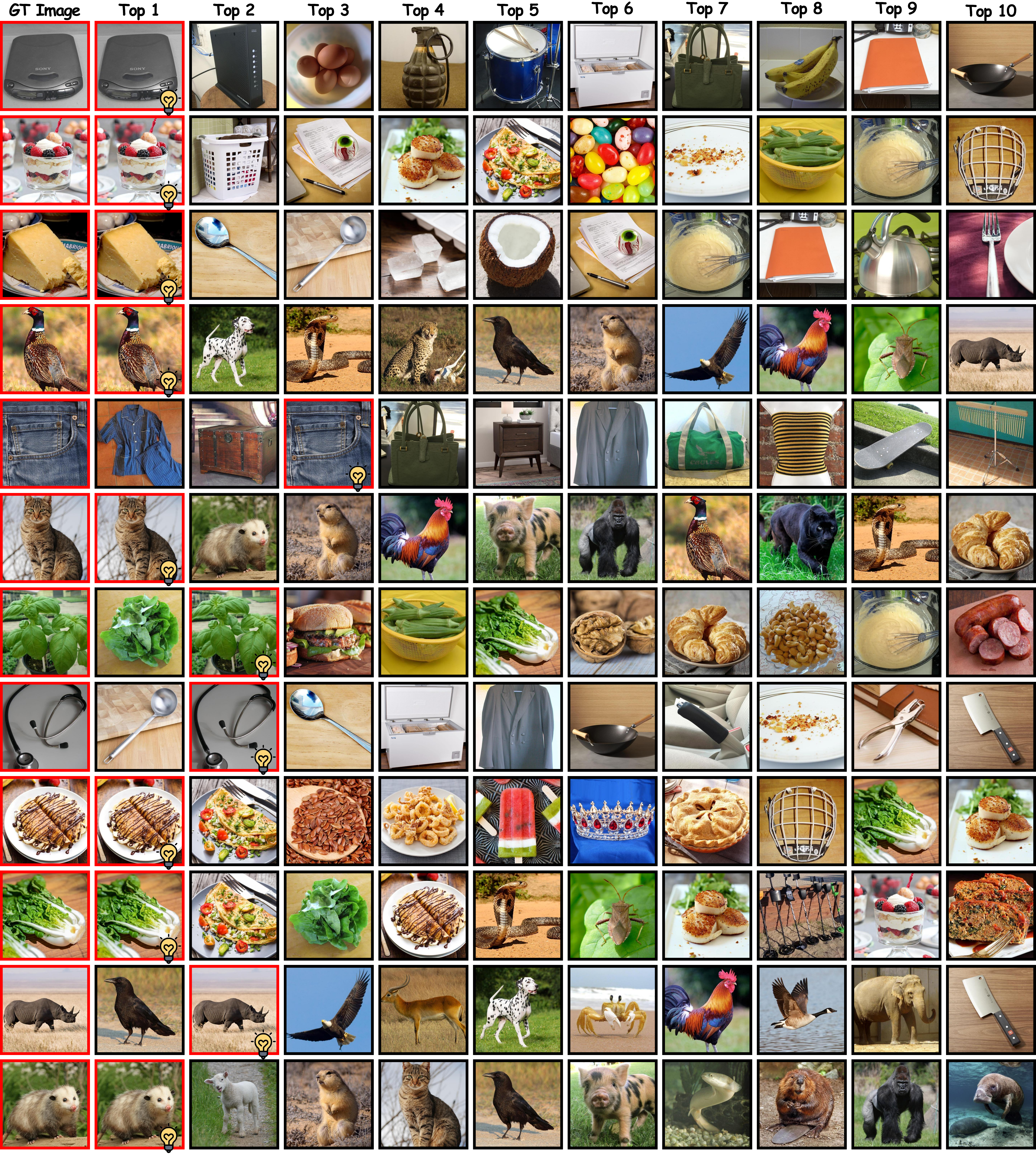}
    \caption{Details of 12 Examples in Top-10 retrieval comparison performance visualization.}
    \label{fig:7}
\end{figure*}
\begin{figure*}[t]
    \centering
    \includegraphics[width=1\linewidth]{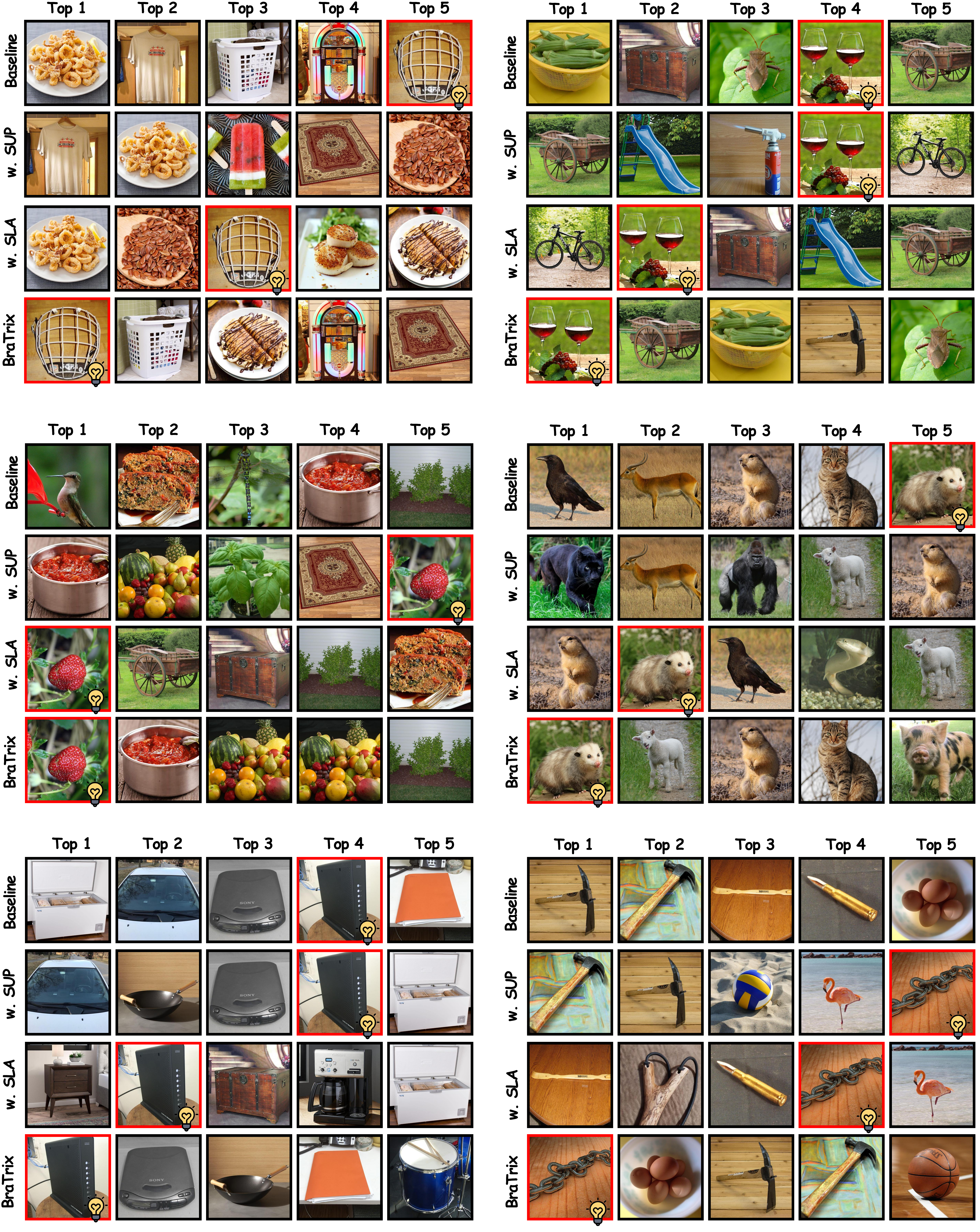}
    \caption{Details of 6 examples in Top-5 retrieval comparison performance visualization in module abltion experiment.}
    \label{fig:8}
\end{figure*}
\begin{figure*}[t]
    \centering
    \includegraphics[width=1\linewidth]{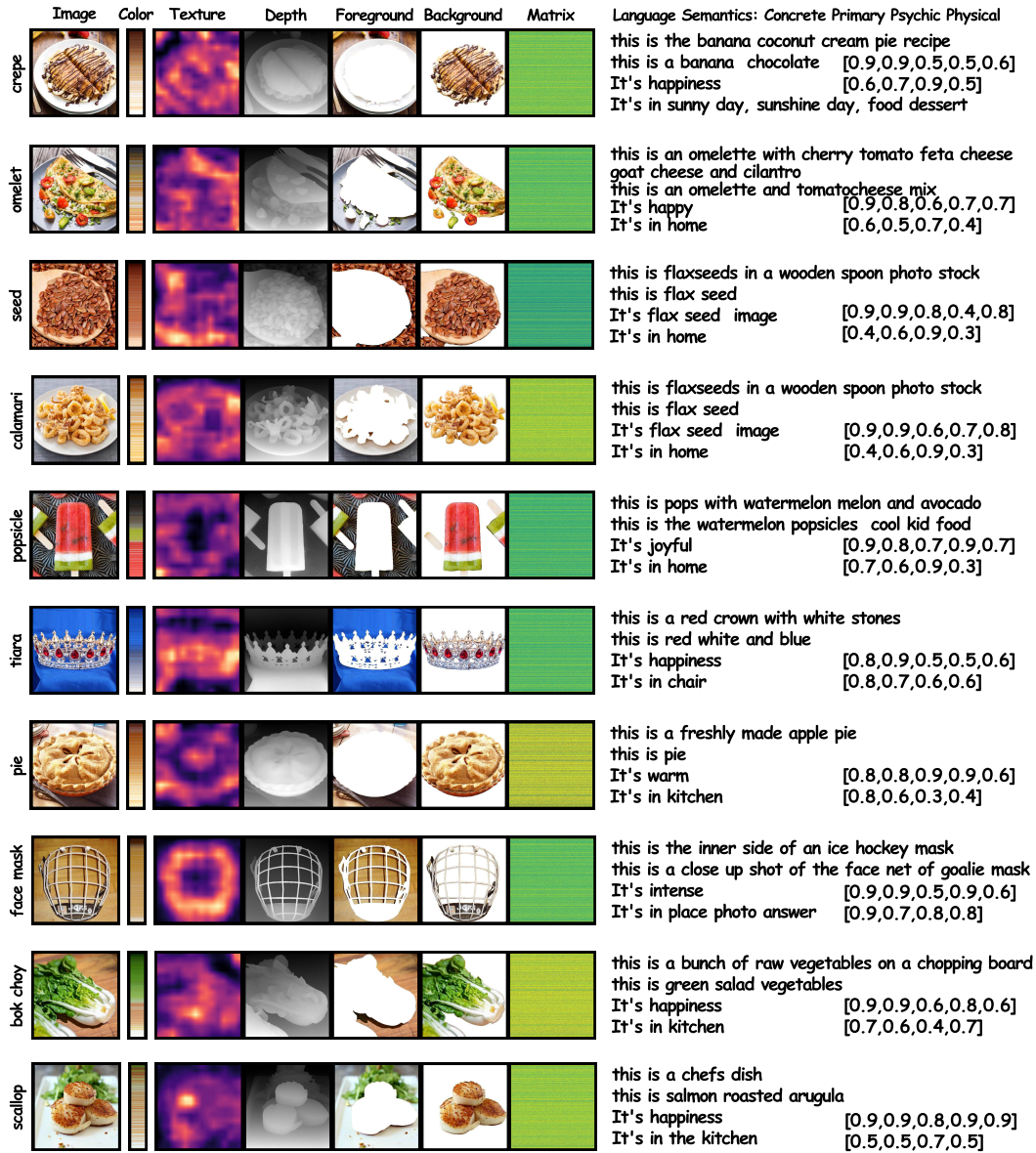}
    \caption{10 detail examples of multimodal semantics, uncertainty perception weight, and Language-EEG matrix visualization. }
    \label{fig:9}
\end{figure*}
\begin{figure*}[t]
    \centering
    \includegraphics[width=1\linewidth]{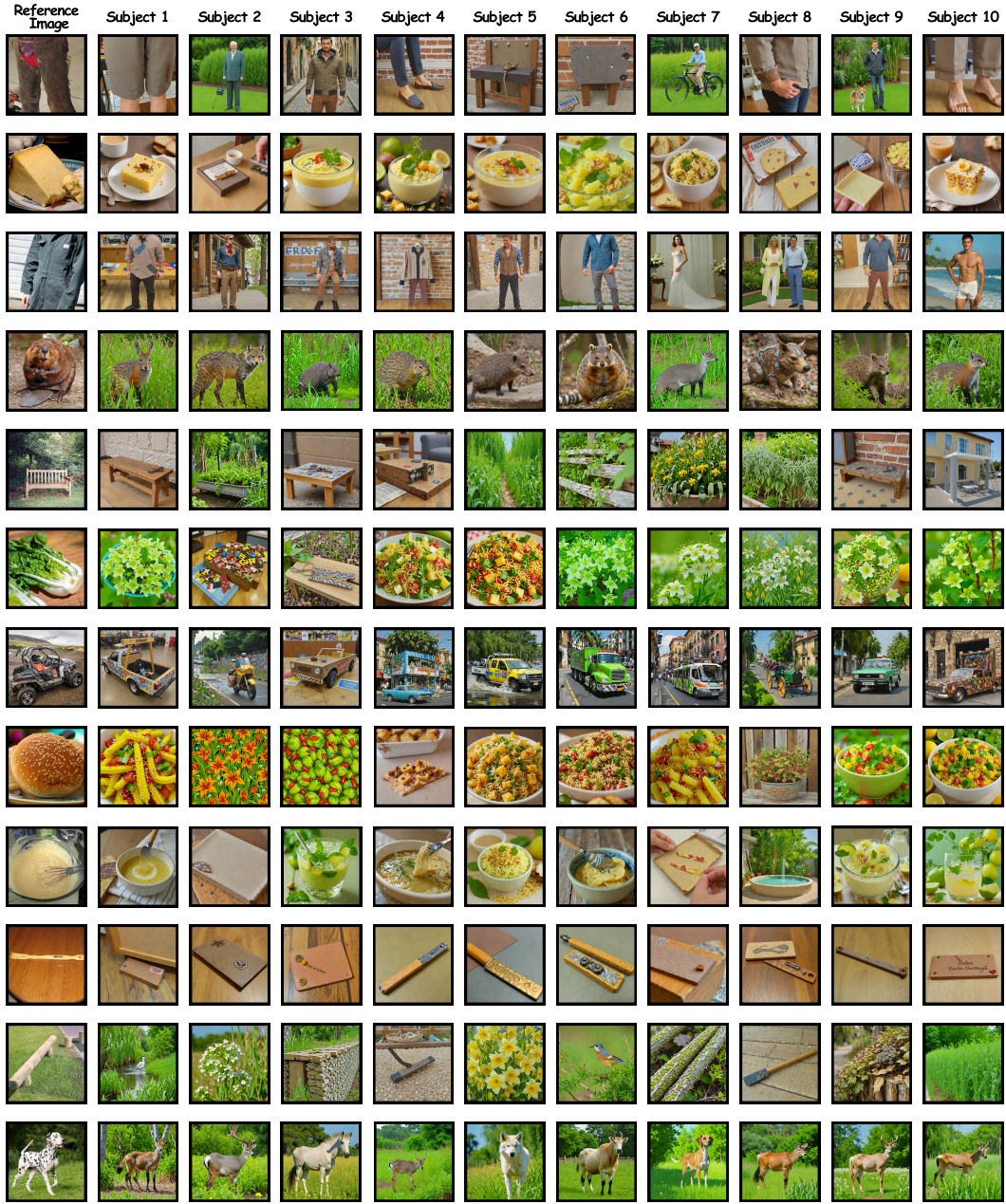}
    \caption{Details of 12 examples in Image generation downstream task on EEG modality across 10 subjects, which demonstrate significant difference between different subjects. }
    \label{fig:10}
\end{figure*}
\begin{figure*}[t]
    \centering
    \includegraphics[width=0.9\linewidth]{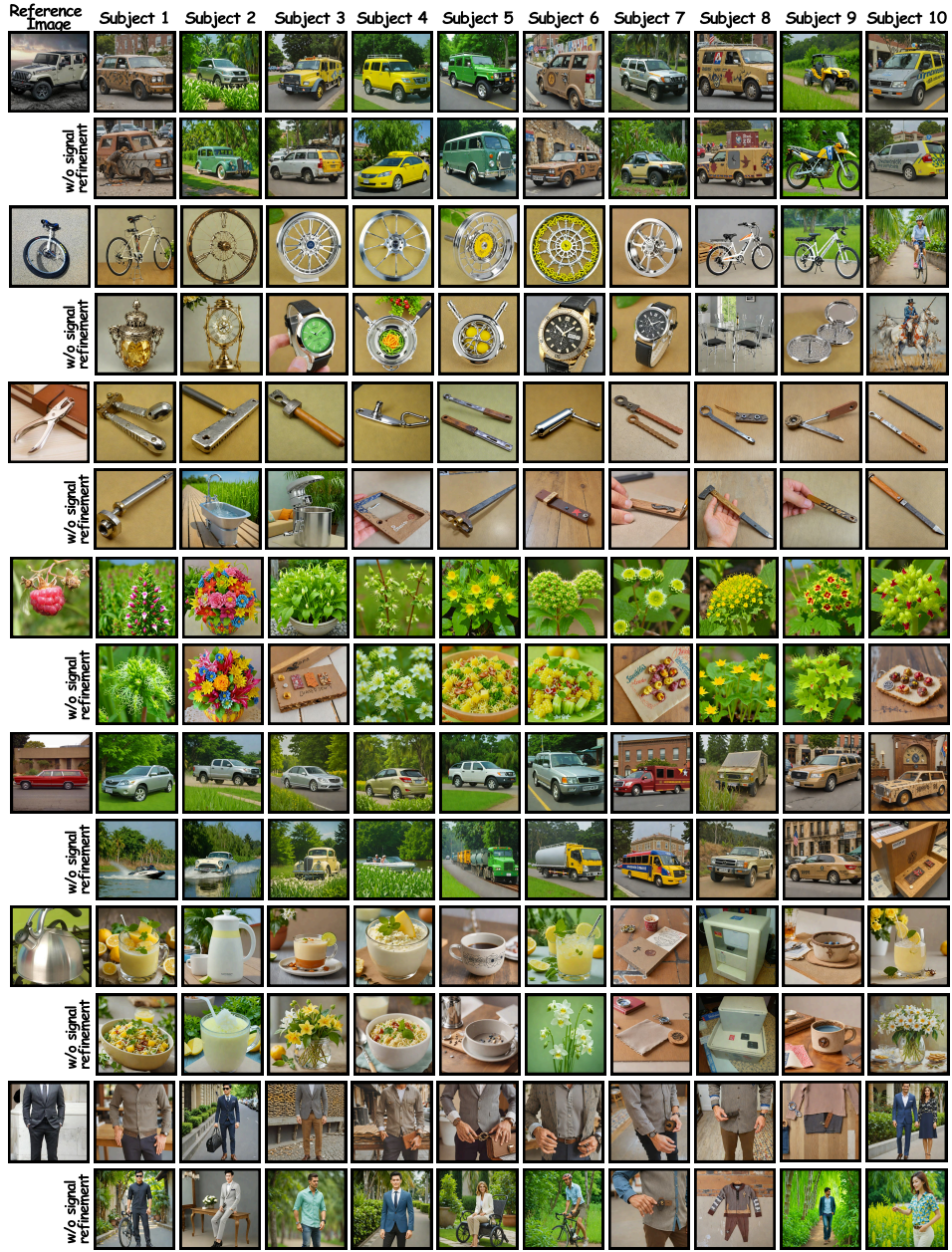}
    \caption{Details of 7 Examples in comparison performance between deep language-anchored visual-brain semantic alignment EEG embeddings and traditional visual-EEG alignment EEG embeddings.}
    \label{fig:11}
\end{figure*}
\begin{figure*}[t]
    \centering
    \includegraphics[width=1\linewidth]{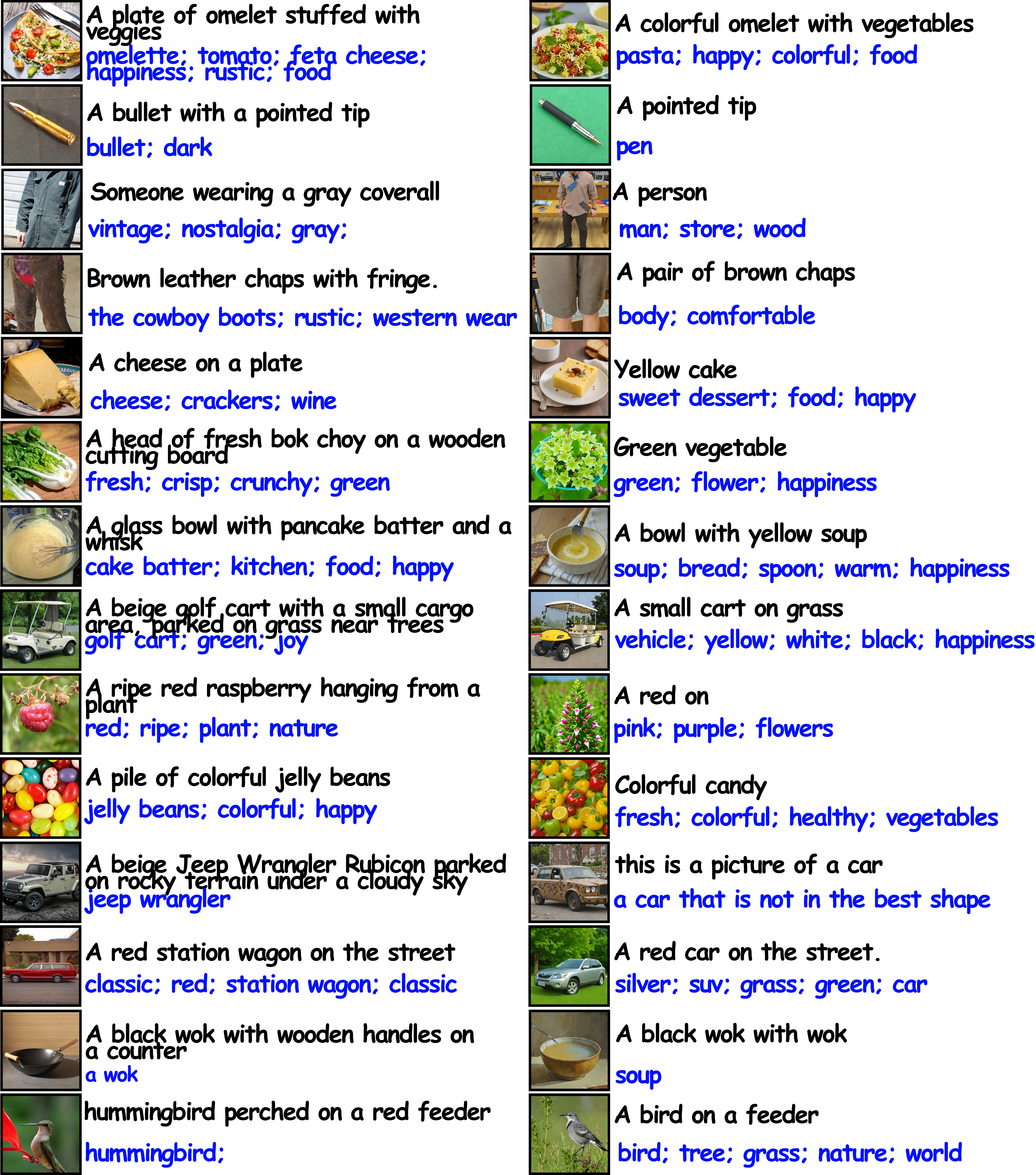}
    \caption{Details of 14 examples in image generation caption downstream tasks on EEG modality in Subject-1. The left column represents the reference images and refined semantic keyboard from per-trained BLIP. The right column represents the generated images and refined semantic keyboard from per-trained BLIP.}
    \label{fig:12}
\end{figure*}
\begin{figure*}[t]
    \centering
    \includegraphics[width=1\linewidth]{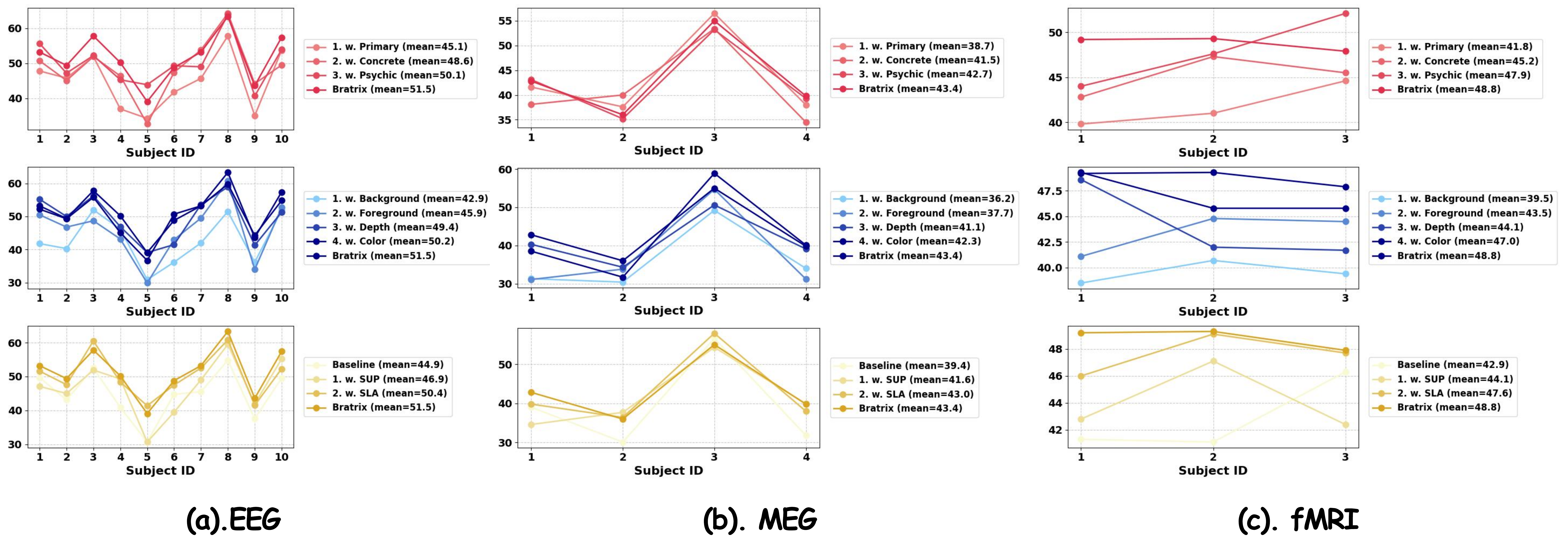}
    \caption{Detail accuracy in 10, 4, and 3 subjects in semantic and module ablation experiments across EEG, MEG, and fMRI modalities.}
    \label{fig:13}
\end{figure*}
\begin{figure*}[t]
    \centering
    \includegraphics[width=1\linewidth]{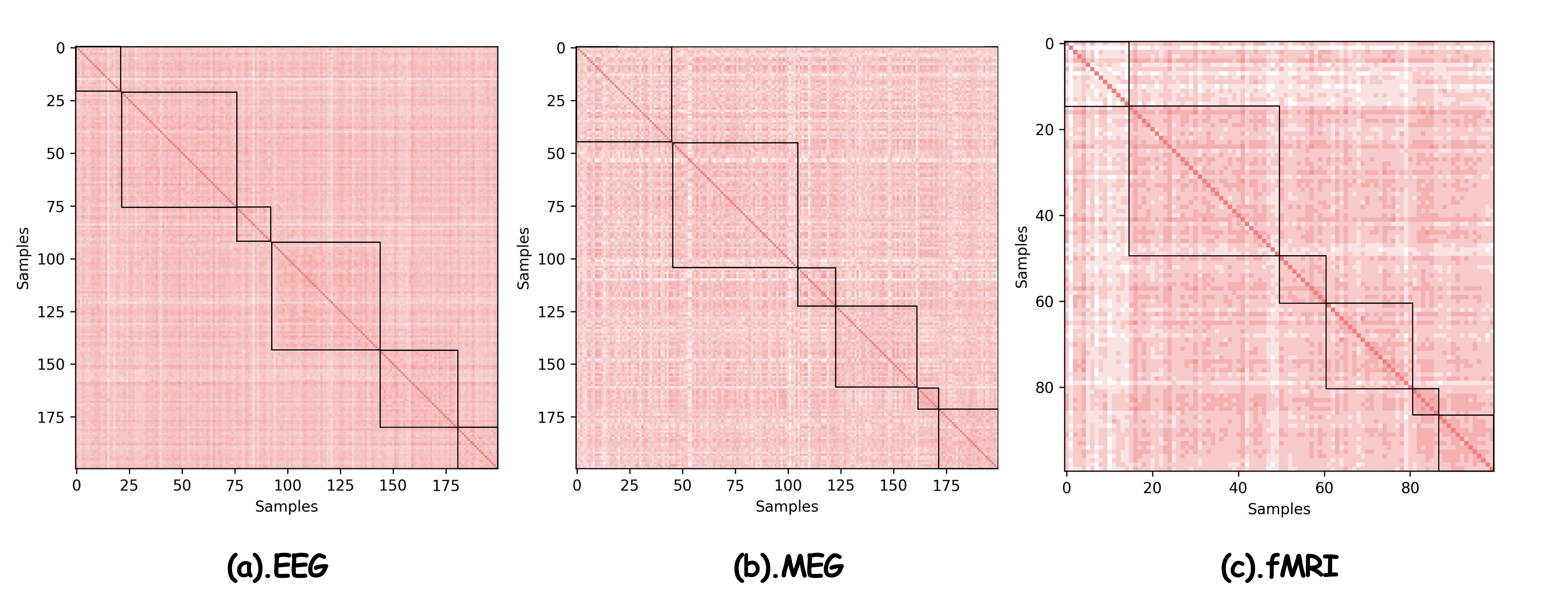}
    \caption{Representation simalarity matrics between brain neural signals and images across categories (Tool, Food, Clothes, Vehicle, Animal, and Others). }
    \label{fig:14}
\end{figure*}
\begin{figure*}[t]
    \centering
    \includegraphics[width=0.4\linewidth]{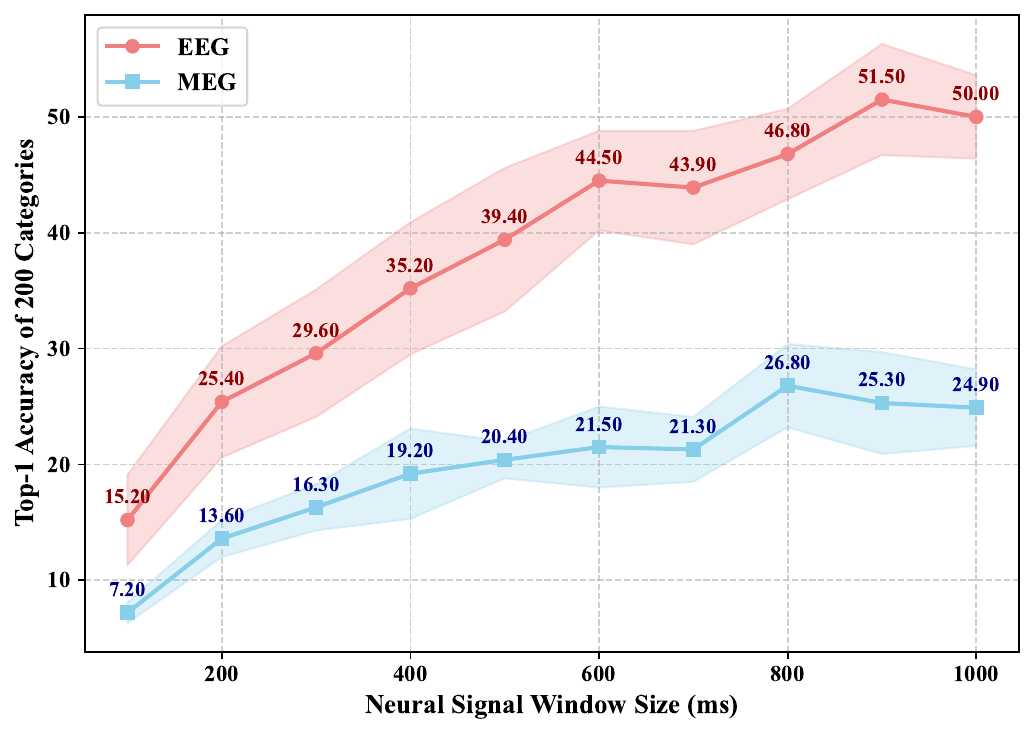}
    \caption{Top-1 Accuracy trends in 200-way in EEG and MEG modalities with different window sizes. }
    \label{fig:15}
\end{figure*}
\section{Detailed Results and Visualization}
\subsection{Quantitative Details in Retrieval Tasks}
\label{app:Quantitative_Retrieval}
Tables~\ref{tab:00} and \ref{tab:01} report image retrieval performance on the EEG modality across subjects under subject-dependent and subject-independent evaluations. Across subjects, EEG retrieval performance is relatively stable, which suggests that the learned EEG representations exhibit reasonable generalization. In the subject-dependent setting, the mean Top-1 accuracies for 4-way, 50-way and 200-way retrieval are 97.9\%, 70.9\% and 51.5\%, respectively. In the subject-independent setting, the mean Top-1 accuracies for 2-way, 50-way and 200-way retrieval are 91.8\%, 32.1\% and 20.5\%, respectively. As expected, performance degrades rapidly as the number of zero-shot retrieval candidates increases, yet Bratrix attains state-of-the-art performance across all reported metrics.

Tables~\ref{tab:02} and \ref{tab:03} show detailed retrieval results for MEG and fMRI modalities. Unlike EEG, MEG and fMRI exhibit pronounced subject-to-subject variability. For example, in the MEG modality the 200-way Top-5 performance of Subject 1 is 38.2\%, whereas Subject 2 reaches 84.8\% in the same setting. Overall retrieval performance also drops sharply as the retrieval set grows: in the MEG subject-dependent experiments performance decreases from 89.0\% (4-way Top-5) to 43.4\% (50-way Top-5) and to 26.8\% (200-way Top-5). Notably, 50-way performance in the subject-independent setting is generally poor across all three modalities; nevertheless, Bratrix remains the state of the art. In contrast, 2-way performance exceeds 80\% for all modalities, indicating that the model retains strong binary discrimination even when multi-class generalization across subjects remains challenging. These results underscore that Bratrix improves alignment performance while subject-specific representational differences remain a limiting factor.
\subsection{Qualitative Details in Retrieval Tasks}
\label{app:Qualitative_Retrieval}
Figures~\ref{fig:7}--\ref{fig:9} present additional qualitative visualizations for comparison and ablation studies. As shown in Fig.~\ref{fig:7}, we randomly visualize 12 examples of Top-10 retrieval results; retrieved samples are typically concentrated within the Top-3 and show strong semantic relevance within the Top-10. As an illustrative case, one “dessert” query returns images that are similar both in low-level appearance, such as a laundry basket with comparable color distribution, and in higher-level semantics, such as jelly beans, omelets, and okra. For animal queries the Top-10 results are predominantly animal images, which suggests that animal-related semantics are represented more uniformly in the neural signal space relative to other categories.

Figure~\ref{fig:8} shows six retrieval examples from the module-level ablation study. Although example-specific variation exists, the addition of the SLA module consistently yields the largest improvement in retrieval precision compared to other module combinations, which empirically supports the effectiveness of the language-anchored alignment design. As more modules are stacked, the semantic coherence among Top-5 samples improves and reaches its best performance when the full Bratrix configuration is used.

We further provide ten detailed examples of multimodal semantic visualizations for image and language modalities in \ref{fig:9}. We also visualize uncertainty weights for image and language semantics as well as the language-anchored EEG semantic matrices; these visuals reveal meaningful differences across images. For instance, the image concept “popsicle” maintains a relatively high image-semantic confidence, while a language-semantic attribute such as “it is in home” receives a low uncertainty weight (around 0.3), which aligns with human interpretation that the location attribute is not strongly supported by the image. Similarly, for “bok choy” the image-semantic confidence is high but the language attribute “it’s happiness” has a low weight (approximately 0.4), again matching intuitive semantic uncertainty.
\subsection{Qualitative Details in Reconstruction Task}
We visualize twelve image reconstruction examples across ten subjects under the EEG modality to analyze subject-level behavior in \ref{fig:10}. The results show substantial inter-subject variability: for example, in one case (second row) a particular subject nearly reconstructs the target object “cheese” while other subjects reconstruct broader food-related content such as noodles, mashed potato, or biscuits. Nevertheless, the majority of reconstructions recover the core semantic category of the target (for example, animal in row four, vehicle in row six, and natural scene in row eleven). Some examples (e.g., row eight) only recover coarse, hierarchical semantics without reproducing fine-grained visual details such as a hamburger appearance. Overall, Bratrix reliably reconstructs the primary semantic content of images from EEG.

We also ran a controlled comparison reported in Fig.~\ref{fig:11}. For each example, the top row shows reconstructions produced after aligning language-anchored deep EEG semantics with CLIP image embeddings, while the bottom row shows reconstructions from a direct alignment between raw EEG embeddings and image embeddings. The language-anchored alignment consistently yields reconstructions that preserve more complete low-level semantics and richer high-level details. For instance, the final example in Fig.~\ref{fig:11} constrains the reconstructed object to “suit” under language-anchored alignment, whereas the direct alignment produces a more generic “person” reconstruction. These comparisons validate the core idea that deep, language-guided semantic refinement materially improves reconstruction fidelity.
\subsection{Qualitative Details in Captioning Task}
We visualize 14 detailed examples of image generation and captioning for Subject 1 in Fig.~\ref{fig:12}. The refined semantic captions produced by Bratrix accurately describe the reference images and correspond closely to the semantics of the reconstructed images. For each reference we generate five candidate reconstructions and select the best; the selected reconstructions and their captions demonstrate consistent alignment with CLIP image representations. This result indicates that, once neural and image embeddings are aligned in a common semantic space, the unified representations are both descriptive of the original stimuli and compatible with off-the-shelf image encoders.

\subsection{Quantitative Details in Ablation Studies}
\label{app:Quantitative_ablation}
We provide detailed ablation results across subjects to assess the contribution of each semantic component and module. As illustrated in Fig.~\ref{fig:13}, subject-to-subject performance differences are substantial, particularly in EEG and MEG. In aggregate, progressive integration of semantic components and modules leads to consistent performance improvements at the individual-subject level across all three modalities. Locally, however, a minority of subjects exhibit slight performance degradation when additional semantics or modules are added. This pattern demonstrates that although multimodal and semantic enhancements are broadly beneficial, their effectiveness can be modulated by subject-specific factors.
\subsection{Comprehensive Multimodal Analyses}
\label{app:comprehensive}
We also visualize representational similarity matrices (RSMs) between brain signals and images across semantic categories in Fig.~\ref{fig:14}. Compared to single-modality brain RSMs, the brain–image RSMs show reduced discriminability, which reflects the intrinsic difficulty of direct RSM comparison across modalities with disparate measurement characteristics. Nevertheless, we observe nontrivial cross-modal structure. This preserved discriminability is attributable in part to the SLA module, where the KL-divergence regularization progressively aligns modality embeddings into a shared semantic latent space.

Finally, we report a temporal window-size study in the subject-dependent setting. We sweep EEG and MEG window sizes from 100 ms to 1000 ms and evaluate 200-way Top-1 retrieval accuracy (fMRI is not included due to the absence of temporal windows). As illustrated in Fig.~\ref{fig:15}, both EEG and MEG accuracies increase steadily as window size grows and reach a plateau around 800–1000 ms, with the largest window achieving the best accuracy. This result demonstrates that temporal information within this interval is critical for accurate decoding in EEG and MEG.
\section{Further methodological insights}
\textbf{Why we perform a cross-matrix computation between language embeddings and visual/brain embeddings. }The cross-matrix computation between embeddings from different modalities, such as visual and neural embeddings, is performed to achieve maximal interaction in a language-anchored semantic space. During training, language embeddings serve as a structured semantic anchor, guiding the alignment of heterogeneous modalities and capturing rich compositional semantics that may not be fully represented in visual or neural features alone. Matrix multiplication explicitly models interactions between each semantic component across modalities, effectively disentangling meaningful correlations from noise and uncertainty in neural recordings. It ensures that during inference (when language input is unavailable) the learned representations retain their semantic structure and cross-modal alignment, because the influence of language during training has already shaped the latent space. In this way, language functions as a semantic reference that enhances feature fusion without weakening multimodal representations, enabling robust and interpretable alignment between brain and visual stimuli.
\section{Societal impacts}
Bratrix, first language-anchored brain–vision alignment framework, has the potential to advance both neuroscience and AI. By revealing how visual and linguistic semantics are jointly represented in the brain, it could enable breakthroughs in neural decoding, BCIs, and cognitive rehabilitation—empowering individuals with speech or motor impairments to communicate via brain-to-language translation or supporting adaptive neurofeedback for attention and memory enhancement. Additionally, using language as a structural anchor may yield more interpretable, human-aligned AI systems that grasp not only visual patterns but their underlying conceptual meanings, bridging perceptual AI and cognitive science.
\section{Ethical statement}
All experiments in this study are conducted on publicly available, anonymized EEG, MEG, and fMRI datasets that comply with their respective institutional ethical approvals and participant consent requirements. No new human or animal data were collected for this research. The Bratrix framework is designed solely for scientific understanding and responsible AI–neuroscience integration; it does not involve any invasive or manipulative neural interventions. We strongly advocate for the ethical, privacy-preserving, and consent-based use of neural data and emphasize that potential downstream applications should adhere to rigorous ethical and legal standards regarding data security and human subject protection.
\begin{sidewaystable*}[t]
\caption{200-way Top-1 and Top-5 image retrieval comparison Performance of Bratrix across 10 subjects on THINGS-EEG2 dataset under subject-dependent and subject-independent setting.}
\centering
\scriptsize
\renewcommand{\arraystretch}{1}
\setlength{\tabcolsep}{3pt}
\begin{tabular}{lrrrrrrrrrrrrrrrrrrrrrrrrrr}
    \toprule
    \multirow{2}{*}{Method} & \multicolumn{2}{c}{Subject 1} & \multicolumn{2}{c}{Subject 2} & \multicolumn{2}{c}{Subject 3} & \multicolumn{2}{c}{Subject 4} & \multicolumn{2}{c}{Subject 5} & \multicolumn{2}{c}{Subject 6} & \multicolumn{2}{c}{Subject 7} & \multicolumn{2}{c}{Subject 8} & \multicolumn{2}{c}{Subject 9} & \multicolumn{2}{c}{Subject 10} & \multicolumn{2}{c}{Average} \\
    \cmidrule(lr){2-3} \cmidrule(lr){4-5} \cmidrule(lr){6-7} \cmidrule(lr){8-9} \cmidrule(lr){10-11} \cmidrule(lr){12-13} \cmidrule(lr){14-15} \cmidrule(lr){16-17} \cmidrule(lr){18-19} \cmidrule(lr){20-21} \cmidrule(lr){22-23}
    & top-1 & top-5 & top-1 & top-5 & top-1 & top-5 & top-1 & top-5 & top-1 & top-5 & top-1 & top-5 & top-1 & top-5 & top-1 & top-5 & top-1 & top-5 & top-1 & top-5 & top-1 & top-5 \\
    \midrule
    \multicolumn{23}{c}{Subject dependent - train and test on one subject} \\
    \midrule
    LSTM \cite{vennerød2021longshorttermmemoryrnn} & 11.3 & 20.4 & 12.7 & 21.8 & 12.9 & 19.6 & 16.8 & 23.2 & 10.7 & 18.2 & 15.2 & 21.7 & 14.1 & 20.0 & 19.3 & 24.0 & 13.1 & 18.7 & 20.1 & 24.5 & 14.6 & 21.0 \\
    ConvNet \cite{radford2021learningtransferablevisualmodels}& 9.7 & 16.7 & 12.1 & 19.2 & 11.2 & 18.1 & 16.7 & 22.6 & 10.1 & 15.8 & 13.2 & 18.4 & 13.6 & 19.6 & 18.2 & 23.1 & 11.8 & 17.5 & 19.7 & 23.0 & 13.4 & 19.1 \\
    EEGNet \cite{Lawhern_2018} & 12.1 & 21.0 & 14.2 & 23.1 & 15.2 & 21.9 & 18.3 & 24.4 & 12.6 & 20.1 & 16.1 & 22.7 & 15.7 & 21.5 & 22.1 & 24.9 & 14.0 & 20.6 & 21.6 & 23.9 & 16.2 & 22.4 \\
    \hline
    MindEyeV2 \cite{scotti2024mindeye2sharedsubjectmodelsenable}& 22.8 & 54.5 & 25.7 & 56.5 & 27.5 & 59.8 & 31.5 & 67.8 & 20.8 & 48.1 & 27.9 & 60.5 & 24.0 & 57.5 & 39.5 & 70.8 & 23.8 & 56.2 & 29.8 & 64.8 & 27.5 & 59.5 \\
    BraVL \cite{du2023decodingvisualneuralrepresentations}& 25.1 & 57.9 & 26.0 & 54.2 & 27.8 & 61.0 & 33.2 & 66.1 & 23.0 & 50.5 & 31.0 & 61.5 & 27.0 & 55.8 & 36.8 & 68.3 & 25.7 & 58.3 & 32.5 & 69.1 & 28.5 & 60.4 \\
    Mb2C \cite{3681292}& 24.5 & 58.2 & 25.3 & 53.5 & 28.1 & 60.8 & 32.8 & 65.8 & 22.7 & 50.1 & 30.4 & 61.3 & 26.6 & 55.4 & 36.3 & 68.0 & 25.4 & 58.1 & 32.1 & 69.0 & 28.1 & 60.1 \\
    NICE \cite{song2024decodingnaturalimageseeg}& 22.3 & 54.2 & 25.2 & 56.1 & 27.2 & 59.4 & 31.1 & 67.5 & 20.3 & 47.8 & 27.4 & 60.2 & 23.6 & 57.1 & 39.1 & 70.6 & 23.4 & 55.9 & 29.4 & 64.6 & 27.1 & 59.2 \\
    ATM-S \cite{NEURIPS2024_ba5f1233}& 26.0 & 61.0 & 23.5 & 55.0 & 24.0 & 60.0 & 30.0 & 62.0 & 13.5 & 45.0 & 22.0 & 50.0 & 28.5 & 60.0 & 36.0 & 71.0 & 23.0 & 53.5 & 28.0 & 64.0 & 25.5 & 58.0 \\
    UBP \cite{11094846}& 30.0 & 58.0 & 28.0 & 61.0 & 35.0 & 66.0 & 33.5 & 60.0 & 23.0 & 50.0 & 31.5 & 57.5 & 32.5 & 61.0 & 44.0 & 70.0 & 34.0 & 55.5 & 36.0 & 64.5 & 33.0 & 59.5 \\
    CogCap \cite{zhang2024cognitioncapturerdecodingvisualstimuli}& 33.0 & 65.0 & 34.0 & 62.0 & 41.0 & 70.0 & 40.0 & 67.5 & 27.0 & 54.0 & 37.0 & 64.0 & 36.0 & 65.0 & 50.0 & 77.0 & 38.0 & 61.5 & 39.5 & 68.5 & 37.5 & 65.5 \\
    ViEEG \cite{liu2025vieeghierarchicalvisualneural}& 31.0 & 68.0 & 35.5 & 64.5 & 39.0 & 71.5 & 49.0 & 78.0 & 29.0 & 59.5 & 45.0 & 75.0 & 40.5 & 73.0 & 53.0 & 81.0 & 38.5 & 72.0 & 44.0 & 79.0 & 40.5 & 72.0 \\
    FLORA \cite{li2025brainflorauncoveringbrainconcept}& 30.0 & 67.0 & 36.0 & 64.0 & 38.5 & 70.5 & 48.0 & 77.0 & 28.0 & 58.0 & 43.5 & 73.0 & 39.0 & 72.0 & 51.5 & 80.0 & 37.5 & 71.0 & 43.0 & 77.5 & 39.5 & 71.0 \\
    \rowcolor{purple!10} Bratrix & 53.2 & 87.7 & 49.4 & 78.3 & 57.8 & 84.4 & 50.2 & 83.6 & 39.1 & 73.5 & 48.8 & 82.1 & 53.2 & 85.4 & 63.3 & 90.5 & 43.6 & 82.5 & 57.4 & 94.5 & 51.5 & 84.5 \\
    
    \midrule
    \multicolumn{23}{c}{Subject independent - leave one subject out for test} \\
    \midrule
    LSTM \cite{vennerød2021longshorttermmemoryrnn}& 6.7 & 15.9 & 8.3 & 17.0 & 7.5 & 14.5 & 9.1 & 18.3 & 6.8 & 13.8 & 8.5 & 16.8 & 7.2 & 14.7 & 9.2 & 18.0 & 6.9 & 15.4 & 8.8 & 17.5 & 7.9 & 16.1 \\
    ConvNet {[}8{]} & 7.1 & 16.4 & 8.8 & 17.9 & 7.6 & 15.0 & 9.4 & 19.0 & 7.3 & 14.4 & 8.7 & 17.4 & 7.5 & 15.3 & 10.0 & 18.9 & 7.0 & 16.1 & 9.0 & 18.2 & 8.2 & 16.9 \\
    EEGNet \cite{Lawhern_2018} & 8.0 & 17.2 & 9.3 & 18.8 & 8.1 & 15.6 & 10.2 & 19.7 & 7.8 & 15.0 & 9.1 & 18.1 & 8.3 & 16.0 & 10.7 & 20.0 & 7.5 & 16.8 & 9.5 & 18.6 & 9.0 & 17.6 \\
    \hline
    MindEyeV2 \cite{scotti2024mindeye2sharedsubjectmodelsenable}& 5.8 & 12.0 & 6.4 & 13.1 & 5.9 & 11.6 & 6.9 & 14.0 & 5.7 & 11.9 & 6.5 & 13.4 & 5.6 & 12.3 & 7.2 & 14.5 & 5.9 & 12.7 & 6.8 & 13.5 & 6.1 & 13.0 \\
    BraVL \cite{du2023decodingvisualneuralrepresentations}& 5.3 & 12.6 & 6.1 & 13.4 & 5.7 & 11.8 & 6.6 & 14.2 & 5.4 & 11.4 & 6.2 & 13.7 & 5.8 & 12.0 & 7.0 & 14.6 & 5.5 & 12.8 & 6.4 & 13.8 & 6.0 & 13.1 \\
    Mb2C \cite{3681292}& 15.5 & 31.1 & 17.0 & 33.7 & 16.2 & 30.0 & 18.9 & 35.1 & 15.8 & 28.5 & 17.4 & 32.9 & 16.6 & 31.0 & 20.1 & 35.9 & 16.0 & 30.4 & 18.5 & 34.3 & 17.2 & 32.3 \\
    NICE \cite{song2024decodingnaturalimageseeg}& 18.4 & 35.4 & 20.3 & 37.9 & 19.1 & 33.7 & 21.9 & 39.5 & 17.7 & 32.0 & 19.9 & 36.3 & 19.0 & 33.9 & 23.1 & 40.9 & 17.9 & 35.1 & 21.2 & 38.7 & 19.9 & 36.3 \\
    ATM-S \cite{NEURIPS2024_ba5f1233}& 16.6 & 33.3 & 18.1 & 35.9 & 17.0 & 31.9 & 19.7 & 37.6 & 16.4 & 30.7 & 17.9 & 34.8 & 17.3 & 32.5 & 20.4 & 38.4 & 16.9 & 32.9 & 19.1 & 36.7 & 17.9 & 34.4 \\
    UBP \cite{11094846}& 15.2 & 30.8 & 16.9 & 29.3 & 16.0 & 28.7 & 18.1 & 31.6 & 14.9 & 27.8 & 16.6 & 30.2 & 15.9 & 28.9 & 17.7 & 32.4 & 15.8 & 28.3 & 16.8 & 31.0 & 16.4 & 29.9 \\
    CogCap \cite{zhang2024cognitioncapturerdecodingvisualstimuli}& 15.1 & 31.4 & 16.1 & 28.9 & 16.7 & 25.4 & 17.4 & 32.0 & 14.7 & 27.6 & 16.9 & 30.9 & 15.6 & 29.4 & 17.9 & 33.1 & 15.4 & 28.7 & 17.1 & 31.6 & 16.3 & 30.3 \\
    ViEEG \cite{liu2025vieeghierarchicalvisualneural}& 18.6 & 47.9 & 20.8 & 36.6 & 18.9 & 37.4 & 13.4 & 36.9 & 15.7 & 33.5 & 21.4 & 46.4 & 20.4 & 43.4 & 22.9 & 47.4 & 18.9 & 34.9 & 19.2 & 50.9 & 19.0 & 41.5 \\
    FLORA \cite{li2025brainflorauncoveringbrainconcept}& 17.5 & 38.7 & 18.7 & 36.4 & 19.8 & 37.6 & 19.3 & 37.2 & 16.6 & 34.4 & 20.3 & 40.1 & 18.4 & 37.4 & 20.9 & 41.2 & 17.9 & 35.7 & 19.9 & 39.7 & 18.9 & 37.9 \\
    \rowcolor{purple!10} Bratrix & 23.5 & 52.5 & 17.5 & 39.0 & 19.0 & 40.5 & 15.5 & 38.5 & 14.5 & 37.5 & 23.5 & 55.0 & 22.0 & 45.5 & 25.5 & 47.5 & 21.5 & 39.5 & 19.5 & 57.5 & 20.5 & 45.3 \\
    
    \bottomrule
\end{tabular}
\label{tab:00}
\end{sidewaystable*}
\begin{sidewaystable*}[t]
\caption{Image retrieval comparison performance of Bratrix across 10 subjects on THINGS-EEG2 dataset under subject-dependent (4-way Top-1 and 50-way Top-1) and subject-independent (2-way Top-1 and 50-way Top-1) setting.}
\centering
\scriptsize
\renewcommand{\arraystretch}{1}
\setlength{\tabcolsep}{3pt}
\begin{tabular}{lrrrrrrrrrrrrrrrrrrrrrrrrrr}
    \toprule
    \multirow{2}{*}{Method} & \multicolumn{2}{c}{Subject 1} & \multicolumn{2}{c}{Subject 2} & \multicolumn{2}{c}{Subject 3} & \multicolumn{2}{c}{Subject 4} & \multicolumn{2}{c}{Subject 5} & \multicolumn{2}{c}{Subject 6} & \multicolumn{2}{c}{Subject 7} & \multicolumn{2}{c}{Subject 8} & \multicolumn{2}{c}{Subject 9} & \multicolumn{2}{c}{Subject 10} & \multicolumn{2}{c}{Average} \\
    \cmidrule(lr){2-3} \cmidrule(lr){4-5} \cmidrule(lr){6-7} \cmidrule(lr){8-9} \cmidrule(lr){10-11} \cmidrule(lr){12-13} \cmidrule(lr){14-15} \cmidrule(lr){16-17} \cmidrule(lr){18-19} \cmidrule(lr){20-21} \cmidrule(lr){22-23}
    \multicolumn{23}{c}{Subject dependent - train and test on one subject} \\
    \midrule
    & 4-w & 50-w & 4-w & 50-w & 4-w & 50-w & 4-w & 50-w & 4-w & 50-w & 4-w & 50-w & 4-w & 50-w & 4-w & 50-w & 4-w & 50-w & 4-w & 50-w & 4-w & 50-w \\
    \midrule
    LSTM \cite{vennerød2021longshorttermmemoryrnn}& 60.1 & 40.2 & 60.9 & 40.7 & 59.3 & 39.1 & 61.4 & 41.6 & 60.8 & 40.9 & 60.4 & 40.6 & 60.6 & 40.8 & 61.7 & 41.1 & 60.2 & 40.4 & 61.9 & 41.5 & 60.6 & 40.7 \\
    ConvNet \cite{radford2021learningtransferablevisualmodels}& 63.8 & 42.1 & 64.6 & 43.7 & 62.9 & 42.8 & 65.4 & 44.9 & 63.1 & 42.6 & 64.1 & 43.9 & 64.8 & 42.3 & 65.7 & 44.6 & 64.3 & 43.5 & 65.9 & 44.8 & 64.5 & 43.5 \\
    EEGNet \cite{Lawhern_2018}& 66.9 & 46.8 & 67.1 & 46.3 & 66.4 & 45.2 & 68.7 & 47.9 & 66.1 & 46.6 & 67.9 & 46.9 & 67.3 & 46.4 & 68.2 & 47.1 & 67.6 & 46.1 & 69.4 & 47.6 & 67.9 & 46.7 \\
    \hline
    MindEyeV2 \cite{scotti2024mindeye2sharedsubjectmodelsenable}& 81.6 & 65.9 & 81.2 & 64.8 & 80.7 & 65.3 & 82.9 & 63.7 & 79.6 & 64.2 & 81.4 & 65.6 & 81.1 & 64.9 & 80.3 & 65.1 & 81.8 & 64.6 & 82.4 & 65.7 & 81.2 & 65.0 \\
    BraVL \cite{du2023decodingvisualneuralrepresentations}& 78.9 & 63.7 & 79.8 & 63.9 & 77.6 & 63.4 & 80.1 & 61.2 & 77.9 & 62.8 & 78.3 & 63.8 & 79.4 & 62.5 & 78.6 & 63.1 & 78.1 & 63.6 & 80.7 & 63.3 & 78.8 & 63.3 \\
    Mb2C \cite{3681292}& 72.6 & 60.2 & 74.8 & 60.9 & 71.3 & 61.7 & 75.1 & 59.6 & 70.9 & 60.5 & 73.2 & 60.1 & 74.6 & 59.3 & 71.8 & 61.9 & 74.9 & 60.8 & 75.8 & 60.4 & 73.0 & 60.6 \\
    NICE \cite{song2024decodingnaturalimageseeg}& 82.7 & 66.9 & 83.5 & 65.8 & 80.1 & 66.2 & 84.6 & 63.9 & 81.8 & 65.7 & 82.3 & 66.5 & 83.1 & 64.7 & 81.4 & 66.8 & 82.9 & 65.3 & 84.9 & 66.1 & 82.9 & 66.0 \\
    ATM-S \cite{NEURIPS2024_ba5f1233}& 80.5 & 64.9 & 81.7 & 64.6 & 78.9 & 64.1 & 82.8 & 62.7 & 79.1 & 63.8 & 80.2 & 64.3 & 81.6 & 62.9 & 78.3 & 64.8 & 80.9 & 63.5 & 82.4 & 64.6 & 80.4 & 64.2 \\
    UBP \cite{11094846}& 76.7 & 61.9 & 77.4 & 61.6 & 75.8 & 62.7 & 78.6 & 59.9 & 76.2 & 60.8 & 77.9 & 61.2 & 77.1 & 60.7 & 76.5 & 61.8 & 77.6 & 61.4 & 78.2 & 61.3 & 77.2 & 61.4 \\
    CogCap \cite{zhang2024cognitioncapturerdecodingvisualstimuli}& 77.5 & 62.9 & 78.6 & 61.8 & 75.1 & 63.6 & 78.9 & 60.7 & 77.8 & 61.9 & 78.4 & 62.5 & 78.1 & 60.3 & 77.6 & 62.8 & 78.7 & 61.2 & 79.1 & 62.1 & 78.0 & 62.1 \\
    ViEEG \cite{liu2025vieeghierarchicalvisualneural}& 90.6 & 67.8 & 91.5 & 67.4 & 88.9 & 68.5 & 93.1 & 63.8 & 87.7 & 65.9 & 90.2 & 67.2 & 91.8 & 64.7 & 88.3 & 67.9 & 90.8 & 65.3 & 91.2 & 67.6 & 90.1 & 66.9 \\
    FLORA \cite{li2025brainflorauncoveringbrainconcept}& 88.5 & 66.7 & 89.1 & 65.9 & 87.6 & 66.1 & 92.8 & 62.6 & 85.9 & 63.2 & 88.8 & 66.4 & 90.5 & 62.1 & 87.4 & 66.8 & 88.1 & 63.9 & 91.6 & 66.3 & 88.7 & 65.7 \\
    \rowcolor{purple!10} Bratrix & 98.6 & 75.8 & 97.9 & 68.7 & 99.2 & 72.9 & 98.1 & 54.8 & 94.7 & 62.8 & 95.3 & 68.2 & 99.6 & 74.6 & 99.8 & 84.9 & 95.9 & 67.9 & 99.1 & 78.3 & 97.9 & 70.9 \\

    \midrule
    \multicolumn{23}{c}{Subject independent - leave one subject out for test} \\
    \midrule
    & 2-w & 50-w & 2-w & 50-w & 2-w & 50-w & 2-w & 50-w & 2-w & 50-w & 2-w & 50-w & 2-w & 50-w & 2-w & 50-w & 2-w & 50-w & 2-w & 50-w & 2-w & 50-w \\
    \midrule
    LSTM \cite{vennerød2021longshorttermmemoryrnn}& 56.9 & 10.8 & 57.2 & 11.6 & 55.7 & 10.3 & 57.1 & 12.9 & 56.4 & 11.8 & 57.8 & 10.1 & 56.1 & 10.9 & 57.5 & 12.4 & 56.3 & 10.7 & 58.6 & 12.1 & 57.0 & 11.3 \\
    ConvNet \cite{radford2021learningtransferablevisualmodels}& 60.7 & 13.9 & 61.2 & 14.8 & 59.8 & 13.6 & 62.5 & 15.7 & 60.3 & 14.2 & 61.9 & 14.1 & 61.6 & 13.4 & 62.1 & 15.2 & 61.4 & 13.8 & 62.2 & 15.4 & 61.5 & 14.3 \\
    EEGNet \cite{Lawhern_2018}& 63.7 & 17.5 & 64.3 & 18.6 & 62.6 & 16.9 & 65.8 & 19.4 & 63.9 & 17.8 & 64.6 & 18.2 & 63.2 & 17.1 & 65.1 & 18.9 & 64.9 & 17.3 & 65.2 & 18.7 & 64.5 & 18.1 \\
    \hline
    MindEyeV2 \cite{scotti2024mindeye2sharedsubjectmodelsenable}& 77.9 & 22.1 & 78.6 & 20.9 & 78.3 & 21.5 & 80.1 & 20.3 & 76.8 & 20.7 & 79.3 & 21.8 & 77.6 & 20.8 & 78.1 & 21.2 & 79.1 & 22.0 & 80.8 & 20.5 & 78.6 & 21.1 \\
    BraVL \cite{du2023decodingvisualneuralrepresentations}& 83.9 & 25.7 & 84.7 & 24.1 & 82.6 & 25.3 & 85.2 & 23.8 & 81.4 & 24.9 & 83.5 & 25.6 & 83.1 & 23.4 & 82.8 & 25.9 & 83.6 & 24.7 & 85.8 & 25.1 & 83.8 & 24.9 \\
    Mb2C \cite{3681292}& 78.5 & 21.7 & 79.1 & 21.4 & 77.8 & 21.9 & 80.6 & 20.6 & 76.3 & 20.1 & 78.9 & 21.2 & 78.2 & 20.4 & 77.5 & 21.6 & 78.7 & 21.8 & 80.4 & 21.3 & 78.4 & 21.2 \\
    NICE \cite{song2024decodingnaturalimageseeg}& 81.9 & 22.8 & 82.4 & 22.5 & 80.7 & 23.6 & 83.8 & 20.9 & 80.2 & 21.7 & 81.6 & 22.1 & 81.3 & 20.3 & 80.9 & 22.4 & 81.8 & 21.2 & 83.5 & 23.8 & 81.9 & 22.3 \\
    ATM-S \cite{NEURIPS2024_ba5f1233}& 78.3 & 20.4 & 79.5 & 20.8 & 77.1 & 21.6 & 81.9 & 18.3 & 78.8 & 19.2 & 79.7 & 20.1 & 79.2 & 18.9 & 77.9 & 20.7 & 78.6 & 19.6 & 80.3 & 21.9 & 78.9 & 20.2 \\
    UBP \cite{11094846}& 75.9 & 17.7 & 76.8 & 17.4 & 74.6 & 18.6 & 77.4 & 15.2 & 75.2 & 16.3 & 76.5 & 17.9 & 77.3 & 16.8 & 75.7 & 17.2 & 76.1 & 17.5 & 77.9 & 17.1 & 76.4 & 17.2 \\
    CogCap \cite{zhang2024cognitioncapturerdecodingvisualstimuli}& 76.8 & 18.3 & 77.5 & 18.9 & 75.9 & 19.7 & 78.5 & 16.4 & 76.6 & 17.2 & 77.2 & 18.6 & 77.9 & 16.7 & 76.4 & 18.1 & 77.8 & 17.9 & 78.2 & 18.4 & 77.3 & 18.1 \\
    ViEEG \cite{liu2025vieeghierarchicalvisualneural}& 88.2 & 26.9 & 89.4 & 25.3 & 87.1 & 26.4 & 91.3 & 23.1 & 87.9 & 25.8 & 88.6 & 26.7 & 89.1 & 24.6 & 87.6 & 26.1 & 88.9 & 24.2 & 91.7 & 26.3 & 88.7 & 25.8 \\
    FLORA \cite{li2025brainflorauncoveringbrainconcept}& 86.2 & 24.1 & 87.3 & 24.8 & 85.4 & 25.6 & 90.7 & 22.9 & 85.8 & 23.5 & 86.9 & 24.3 & 87.6 & 22.4 & 85.2 & 24.7 & 86.5 & 23.1 & 90.1 & 24.9 & 87.0 & 24.1 \\
    
    \rowcolor{purple!10} Bratrix & 93.2 & 34.1 & 90.8 & 39.2 & 91.5 & 30.8 & 89.7 & 30.6 & 88.2 & 27.9 & 93.6 & 30.1 & 93.4 & 33.8 & 91.1 & 32.9 & 91.8 & 25.1 & 94.2 & 36.7 & 91.8 & 32.1 \\
\bottomrule
\end{tabular}
\label{tab:01}
\end{sidewaystable*}
\begin{sidewaystable*}[t]
\caption{200-way Top-1 and Top-5 image retrieval comparison Performance of Bratrix across 10 subjects on THINGS-MEG and THINGS-fMRI dataset under subject-dependent and subject-independent setting.}
\centering
\scriptsize
\renewcommand{\arraystretch}{1} 
\setlength{\tabcolsep}{3pt}
  \begin{tabular}{lrrrrrrrrrr|rrrrrrrr}
    \toprule
    & \multicolumn{10}{c}{MEG Dataset} & \multicolumn{8}{c}{fMRI Dataset}\\
    \midrule
    \multirow{2}{*}{Method} & \multicolumn{2}{c}{Subject 1} & \multicolumn{2}{c}{Subject 2} & \multicolumn{2}{c}{Subject 3} & \multicolumn{2}{c}{Subject 4} & \multicolumn{2}{c}{Average} & \multicolumn{2}{c}{Subject 1} & \multicolumn{2}{c}{Subject 2} & \multicolumn{2}{c}{Subject 3}  & \multicolumn{2}{c}{Average} \\
    \cmidrule(lr){2-3} \cmidrule(lr){4-5} \cmidrule(lr){6-7} \cmidrule(lr){8-9} \cmidrule(lr){10-11} \cmidrule(lr){12-13} \cmidrule(lr){14-15} \cmidrule(lr){16-17} \cmidrule(lr){18-19} 
    & top-1 & top-5 & top-1 & top-5 & top-1 & top-5 & top-1 & top-5 & top-1 & top-5 & top-1 & top-5 & top-1 & top-5 & top-1 & top-5 & top-1 & top-5 \\
    \midrule
    \multicolumn{19}{c}{Subject dependent - train and test on one subject} \\
    \midrule
    LSTM \cite{vennerød2021longshorttermmemoryrnn}& 9.0 & 14.2 & 10.1 & 18.3 & 8.5 & 20.0 & 7.8 & 15.0 & 9.4 & 18.6 & 13.2 & 21.5 & 9.0 & 20.1 & 8.3 & 22.5 & 9.4 & 23.0 \\
    ConvNet \cite{radford2021learningtransferablevisualmodels}& 10.8 & 16.5 & 12.3 & 20.7 & 9.7 & 22.8 & 9.9 & 17.6 & 11.1 & 20.3 & 15.7 & 24.1 & 11.3 & 22.0 & 10.5 & 24.6 & 12.8 & 25.0 \\
    EEGNet \cite{Lawhern_2018} & 12.6 & 18.4 & 14.9 & 23.5 & 11.2 & 25.7 & 11.7 & 19.8 & 13.0 & 23.9 & 17.3 & 27.9 & 13.1 & 25.4 & 11.2 & 28.0 & 14.9 & 29.5 \\
    \hline
    MindEyeV2 \cite{scotti2024mindeye2sharedsubjectmodelsenable}& 4.7 & 30.4 & 21.9 & 62.5 & 16.3 & 48.0 & 6.8 & 34.9 & 13.5 & 49.8 & 21.2 & 70.1 & 13.6 & 59.0 & 10.5 & 54.7 & 15.0 & 58.2 \\
    BraVL \cite{du2023decodingvisualneuralrepresentations}& 7.2 & 29.1 & 23.8 & 61.6 & 18.6 & 46.9 & 9.4 & 33.7 & 15.8 & 48.0 & 22.1 & 68.3 & 14.5 & 57.9 & 11.8 & 52.9 & 16.1 & 56.4 \\ 
    Mb2C \cite{3681292}& 11.0 & 28.4 & 22.5 & 50.6 & 20.4 & 47.8 & 13.5 & 31.0 & 19.2 & 44.7 & 25.0 & 60.2 & 18.6 & 51.3 & 13.8 & 52.7 & 20.6 & 55.9 \\
    NICE \cite{song2024decodingnaturalimageseeg}& 12.2 & 30.5 & 24.8 & 54.1 & 21.9 & 49.3 & 14.1 & 33.7 & 21.6 & 46.8 & 27.4 & 63.7 & 22.4 & 53.9 & 16.1 & 55.5 & 22.2 & 57.8 \\
    ATM-S \cite{NEURIPS2024_ba5f1233}& 10.9 & 27.0 & 23.0 & 49.2 & 19.4 & 45.9 & 12.8 & 30.5 & 18.7 & 43.1 & 29.9 & 58.9 & 23.5 & 50.1 & 18.7 & 48.8 & 25.6 & 51.9 \\
    UBP \cite{11094846}& 15.0 & 38.0 & 46.0 & 80.5 & 27.3 & 59.0 & 18.5 & 43.5 & 26.7 & 55.2 & 40.6 & 70.3 & 32.4 & 55.1 & 33.8 & 62.2 & 28.7 & 59.3 \\
    CogCap \cite{zhang2024cognitioncapturerdecodingvisualstimuli}& 13.5 & 33.9 & 31.8 & 71.3 & 24.0 & 53.8 & 14.9 & 35.6 & 22.7 & 48.9 & 37.4 & 66.7 & 33.9 & 60.2 & 26.9 & 58.3 & 32.6 & 62.5 \\
    ViEEG \cite{liu2025vieeghierarchicalvisualneural}& 14.6 & 36.4 & 42.7 & 80.0 & 27.0 & 58.9 & 16.8 & 38.0 & 25.0 & 54.0 & 43.5 & 85.0 & 36.1 & 82.0 & 23.9 & 68.0 & 35.8 & 78.0 \\
    FLORA \cite{li2025brainflorauncoveringbrainconcept}& 13.9 & 35.2 & 41.1 & 78.4 & 26.1 & 57.3 & 15.9 & 36.9 & 24.5 & 52.6 & 46.2 & 83.8 & 39.0 & 80.7 & 26.7 & 66.9 & 38.0 & 76.5 \\
    \rowcolor{purple!10} Bratrix & 15.3 & 38.2 & 46.5 & 84.8 & 28.2 & 61.7 & 17.8 & 39.2 & 26.8 & 56.0 & 54.2 & 87.8 & 46.8 & 84.2 & 32.2 & 70.3 & 44.5 & 80.6 \\

    \midrule
    \multicolumn{19}{c}{Subject independent - leave one subject out for test} \\
    \midrule
    LSTM \cite{vennerød2021longshorttermmemoryrnn}& 1.1 & 5.6 & 2.0 & 6.2 & 2.4 & 5.1 & 2.3 & 6.7 & 1.8 & 5.3 & 2.5 & 6.1 & 1.7 & 5.4 & 2.2 & 6.0 & 1.9 & 5.8 \\
    ConvNet \cite{radford2021learningtransferablevisualmodels}& 1.4 & 6.0 & 2.2 & 6.8 & 1.9 & 5.6 & 2.7 & 7.1 & 2.0 & 5.9 & 2.5 & 6.5 & 1.8 & 6.2 & 2.3 & 6.8 & 2.0 & 6.0 \\
    EEGNet \cite{Lawhern_2018} & 1.6 & 6.3 & 2.4 & 7.2 & 2.0 & 6.0 & 3.0 & 7.5 & 2.2 & 6.1 & 1.8 & 7.0 & 1.9 & 6.5 & 2.5 & 7.2 & 2.1 & 6.4 \\
    \hline
    MindEyeV2 \cite{scotti2024mindeye2sharedsubjectmodelsenable}& 1.9 & 4.3 & 2.0 & 4.5 & 2.1 & 4.2 & 2.2 & 4.8 & 1.8 & 4.1 & 2.3 & 5.0 & 2.0 & 4.4 & 2.4 & 4.9 & 2.1 & 4.6  \\
    BraVL \cite{du2023decodingvisualneuralrepresentations}& 1.8 & 4.2 & 2.1 & 4.7 & 1.9 & 4.1 & 2.3 & 5.0 & 2.0 & 4.3 & 2.4 & 5.1 & 2.1 & 4.5 & 2.3 & 4.9 & 2.2 & 4.6 \\
    Mb2C \cite{3681292}& 2.9 & 6.5 & 3.2 & 7.0 & 3.0 & 6.2 & 3.5 & 7.3 & 3.3 & 6.7 & 3.6 & 7.5 & 3.2 & 6.8 & 3.5 & 7.2 & 3.4 & 6.9 \\
    NICE \cite{song2024decodingnaturalimageseeg}& 3.3 & 7.2 & 3.7 & 7.8 & 3.5 & 6.9 & 4.0 & 8.1 & 3.8 & 7.5 & 4.2 & 8.4 & 3.9 & 7.7 & 4.1 & 8.0 & 4.0 & 7.6 \\
    ATM-S \cite{NEURIPS2024_ba5f1233}& 3.1 & 6.8 & 3.5 & 7.5 & 3.2 & 6.5 & 3.7 & 7.7 & 3.4 & 7.0 & 3.9 & 8.0 & 3.6 & 7.2 & 3.8 & 7.6 & 3.7 & 7.1 \\
    UBP \cite{11094846}& 2.0 & 5.7 & 1.5 & 17.2 & 2.7 & 10.5 & 2.5 & 8.0 & 2.2 & 10.4 & 4.0 & 7.9 & 3.7 & 7.1 & 3.9 & 7.5 & 3.6 & 7.2 \\
    CogCap \cite{zhang2024cognitioncapturerdecodingvisualstimuli}& 4.2 & 9.5 & 4.8 & 10.2 & 4.5 & 9.2 & 5.1 & 10.8 & 4.7 & 9.8 & 5.3 & 11.0 & 4.9 & 10.0 & 5.2 & 10.7 & 5.0 & 10.1 \\
    ViEEG \cite{liu2025vieeghierarchicalvisualneural}& 3.1 & 8.7 & 3.5 & 9.5 & 3.3 & 8.2 & 3.8 & 9.8 & 3.5 & 8.6 & 4.0 & 9.9 & 3.6 & 8.8 & 3.9 & 9.7 & 3.7 & 9.0 \\
    FLORA \cite{li2025brainflorauncoveringbrainconcept}& 3.0 & 8.4 & 3.4 & 9.1 & 3.2 & 7.9 & 3.7 & 9.4 & 3.4 & 8.2 & 3.9 & 9.6 & 3.5 & 8.5 & 3.8 & 9.3 & 3.6 & 8.7 \\
    \rowcolor{purple!10} Bratrix & 4.8 & 12.7 & 6.2 & 17.8 & 4.3 & 15.2 & 4.1 & 12.3 & 4.9 & 14.5 & 4.7 & 13.2 & 4.6 & 12.8 & 3.7 & 11.3 & 4.2 & 12.3 \\
\bottomrule
\end{tabular}
\label{tab:02}
\end{sidewaystable*}
\begin{sidewaystable*}[t]
\caption{Image retrieval comparison Performance of Bratrix across 10 subjects on THINGS-MEG and THINGS-fMRI dataset under subject-dependent (4-way Top-1 and 50-way Top-1) and subject-independent (2-way Top-1 and 50-way Top-1) setting.}
\centering
\scriptsize
\renewcommand{\arraystretch}{1} 
\setlength{\tabcolsep}{3pt}
  \begin{tabular}{lrrrrrrrrrr|rrrrrrrr}
    \toprule
    & \multicolumn{10}{c}{MEG Dataset} & \multicolumn{8}{c}{fMRI Dataset}\\
    \midrule
    \multirow{2}{*}{Method} & \multicolumn{2}{c}{Subject 1} & \multicolumn{2}{c}{Subject 2} & \multicolumn{2}{c}{Subject 3} & \multicolumn{2}{c}{Subject 4} & \multicolumn{2}{c}{Average} & \multicolumn{2}{c}{Subject 1} & \multicolumn{2}{c}{Subject 2} & \multicolumn{2}{c}{Subject 3}  & \multicolumn{2}{c}{Average} \\
    \cmidrule(lr){2-3} \cmidrule(lr){4-5} \cmidrule(lr){6-7} \cmidrule(lr){8-9} \cmidrule(lr){10-11} \cmidrule(lr){12-13} \cmidrule(lr){14-15} \cmidrule(lr){16-17} \cmidrule(lr){18-19} 
    \multicolumn{19}{c}{Subject dependent - train and test on one subject} \\
    \midrule
    & 4-w & 50-w & 4-w & 50-w & 4-w & 50-w & 4-w & 50-w & 4-w & 50-w & 4-w & 50-w & 4-w & 50-w & 4-w & 50-w & 4-w & 50-w\\
   \midrule
    LSTM \cite{vennerød2021longshorttermmemoryrnn}& 47.6 & 8.5 & 53.2 & 11.2 & 49.8 & 13.9 & 48.3 & 9.7 & 49.7 & 10.8 & 51.5 & 16.8 & 50.2 & 14.3 & 47.9 & 12.6 & 49.9 & 14.6 \\
    ConvNet \cite{radford2021learningtransferablevisualmodels}& 49.2 & 10.1 & 55.7 & 13.6 & 52.3 & 16.4 & 50.9 & 11.8 & 52.0 & 13.0 & 54.1 & 19.2 & 52.8 & 16.9 & 50.5 & 15.0 & 52.5 & 17.0 \\
    EEGNet  \cite{Lawhern_2018}  & 51.7 & 11.6 & 57.9 & 14.9 & 54.6 & 17.8 & 53.2 & 13.4 & 54.4 & 14.4 & 56.3 & 20.7 & 55.1 & 18.4 & 52.9 & 16.7 & 54.8 & 18.6 \\
    \hline
    MindEyeV2 \cite{scotti2024mindeye2sharedsubjectmodelsenable}& 52.3 & 12.8 & 58.6 & 15.7 & 54.9 & 18.2 & 53.1 & 14.5 & 54.7 & 15.3 & 56.4 & 21.3 & 55.8 & 19.6 & 52.8 & 17.4 & 55.0 & 19.4 \\
    BraVL \cite{du2023decodingvisualneuralrepresentations}& 55.7 & 14.2 & 62.9 & 17.3 & 58.4 & 20.5 & 56.8 & 16.1 & 58.5 & 17.0 & 60.1 & 23.8 & 59.3 & 21.9 & 56.3 & 19.2 & 58.6 & 21.6 \\
    Mb2C \cite{3681292}& 59.1 & 16.5 & 67.2 & 19.8 & 62.3 & 23.7 & 60.5 & 18.6 & 62.3 & 19.7 & 64.7 & 26.5 & 63.6 & 24.2 & 59.7 & 21.8 & 62.7 & 24.2 \\
    NICE \cite{song2024decodingnaturalimageseeg}& 63.5 & 18.3 & 71.5 & 22.4 & 66.8 & 26.9 & 64.2 & 20.9 & 66.5 & 22.1 & 69.3 & 29.7 & 67.8 & 26.8 & 63.2 & 24.5 & 66.8 & 27.0 \\
    ATM-S \cite{NEURIPS2024_ba5f1233}& 67.8 & 20.6 & 75.8 & 24.9 & 71.2 & 30.1 & 68.4 & 23.3 & 70.8 & 24.7 & 73.9 & 32.8 & 72.1 & 29.5 & 67.1 & 27.3 & 71.0 & 29.9 \\
    UBP \cite{11094846}& 61.2 & 23.1 & 80.3 & 28.5 & 75.6 & 34.8 & 72.9 & 25.7 & 72.5 & 28.0 & 78.5 & 36.2 & 76.4 & 32.7 & 71.5 & 30.4 & 75.5 & 33.1 \\
    CogCap \cite{zhang2024cognitioncapturerdecodingvisualstimuli}& 65.6 & 26.4 & 85.7 & 31.8 & 80.1 & 39.2 & 77.3 & 28.9 & 77.2 & 31.6 & 83.2 & 39.5 & 80.7 & 35.9 & 75.8 & 33.7 & 79.9 & 36.4 \\
    ViEEG \cite{liu2025vieeghierarchicalvisualneural}& 69.3 & 29.8 & 90.4 & 34.2 & 84.7 & 42.6 & 81.6 & 32.4 & 81.5 & 34.8 & 88.6 & 43.9 & 85.2 & 39.4 & 80.2 & 36.8 & 84.7 & 40.0 \\
    FLORA \cite{li2025brainflorauncoveringbrainconcept}& 63.7 & 33.2 & 94.6 & 36.7 & 89.2 & 47.3 & 86.1 & 35.8 & 83.4 & 38.3 & 93.4 & 48.6 & 89.7 & 43.1 & 84.6 & 40.2 & 89.2 & 44.0 \\
    \rowcolor{purple!10} Bratrix & 72.0 & 41.6 & 97.8 & 37.6 & 93.4 & 56.5 & 92.7 & 38.0 & 89.0 & 43.4 & 93.3 & 49.2 & 93.1 & 49.3 & 92.0 & 47.9 & 92.8 & 48.8 \\
    \midrule
    \multicolumn{19}{c}{Subject independent - leave one subject out for test} \\    
    \midrule
    & 2-w & 50-w & 2-w & 50-w & 2-w & 50-w & 2-w & 50-w & 2-w & 50-w & 2-w & 50-w & 2-w & 50-w & 2-w & 50-w & 2-w & 50-w \\
    \midrule
    LSTM \cite{vennerød2021longshorttermmemoryrnn}& 38.2 & 2.1 & 42.5 & 1.8 & 36.7 & 2.5 & 40.1 & 1.5 & 39.4 & 2.0 & 37.6 & 1.2 & 41.8 & 2.3 & 35.9 & 1.9 & 38.4 & 1.8 \\
    ConvNet \cite{radford2021learningtransferablevisualmodels}& 40.5 & 2.3 & 43.8 & 2.0 & 39.1 & 2.7 & 42.3 & 1.7 & 41.4 & 2.2 & 39.8 & 1.4 & 43.2 & 2.5 & 37.2 & 2.1 & 40.1 & 2.0 \\
    EEGNet \cite{Lawhern_2018} & 42.7 & 2.5 & 44.9 & 2.2 & 41.5 & 2.9 & 44.6 & 1.9 & 43.4 & 2.4 & 42.1 & 1.6 & 44.5 & 2.7 & 39.5 & 2.3 & 42.0 & 2.2 \\
    \hline
    MindEyeV2 \cite{scotti2024mindeye2sharedsubjectmodelsenable}& 41.3 & 5.2 & 44.1 & 7.7 & 42.8 & 6.5 & 40.5 & 4.3 & 42.2 & 5.9 & 39.8 & 3.9 & 40.9 & 6.2 & 39.1 & 3.3 & 39.9 & 4.5 \\
    BraVL \cite{du2023decodingvisualneuralrepresentations}& 42.7 & 5.5 & 45.8 & 8.2 & 44.5 & 6.9 & 41.9 & 4.7 & 43.7 & 6.3 & 41.1 & 4.2 & 42.2 & 6.6 & 40.4 & 3.6 & 41.2 & 4.8 \\
    Mb2C \cite{3681292}& 46.2 & 6.1 & 49.3 & 9.1 & 47.9 & 7.6 & 45.4 & 5.3 & 47.2 & 7.0 & 44.6 & 4.8 & 45.8 & 7.3 & 43.9 & 4.1 & 44.8 & 5.4 \\
    NICE \cite{song2024decodingnaturalimageseeg}& 50.6 & 6.8 & 53.7 & 10.3 & 52.2 & 8.4 & 49.8 & 6.0 & 51.6 & 7.9 & 48.9 & 5.5 & 50.1 & 8.1 & 48.2 & 4.7 & 49.1 & 6.1 \\
    ATM-S \cite{NEURIPS2024_ba5f1233}& 54.3 & 7.4 & 57.6 & 11.2 & 56.1 & 9.2 & 53.5 & 6.6 & 55.4 & 8.6 & 52.7 & 6.1 & 53.9 & 8.8 & 51.8 & 5.3 & 52.8 & 6.7 \\
    UBP \cite{11094846}& 58.8 & 8.0 & 62.2 & 12.1 & 60.7 & 9.9 & 57.9 & 7.2 & 59.9 & 9.3 & 56.8 & 6.6 & 58.1 & 9.5 & 55.7 & 5.8 & 56.9 & 7.3 \\
    CogCap \cite{zhang2024cognitioncapturerdecodingvisualstimuli}& 63.5 & 8.5 & 67.1 & 12.9 & 65.5 & 10.6 & 62.6 & 7.7 & 64.7 & 9.9 & 61.5 & 7.1 & 62.8 & 10.2 & 60.3 & 6.3 & 61.5 & 7.9 \\
    ViEEG \cite{liu2025vieeghierarchicalvisualneural}& 67.2 & 8.8 & 71.4 & 13.3 & 69.8 & 11.0 & 66.3 & 7.9 & 68.7 & 10.3 & 65.2 & 7.3 & 66.6 & 10.5 & 63.8 & 6.5 & 65.2 & 8.1 \\
    FLORA \cite{li2025brainflorauncoveringbrainconcept}& 70.4 & 9.0 & 74.6 & 13.6 & 72.9 & 11.1 & 69.5 & 7.8 & 71.9 & 10.4 & 68.3 & 7.2 & 69.7 & 10.6 & 66.9 & 6.4 & 68.3 & 8.1 \\
    \rowcolor{purple!10} Bratrix & 75.6 & 10.1 & 86.9 & 14.8 & 84.7 & 12.2 & 74.0 & 8.7 & 80.3 & 11.5 & 71.2 & 7.2 & 74.5 & 12.5 & 64.6 & 6.9 & 70.1 & 8.9 \\
\bottomrule
\end{tabular}
\label{tab:03}
\end{sidewaystable*}

\end{document}